%% file: main.tex
\documentclass[lettersize,journal]{IEEEtran}
\usepackage[utf8]{inputenc} 
\usepackage[T1]{fontenc}    
\usepackage{url}            
\usepackage{booktabs}       
\usepackage{amsfonts}       
\usepackage{nicefrac}       
\usepackage{microtype}      
\usepackage{color}
\usepackage{subfigure}
\usepackage{caption}
\usepackage{tikz}
\usepackage{pgfplots}
\pgfplotsset{compat=1.8}
\usepgfplotslibrary{statistics, groupplots}
\usepackage{paralist}
\usepackage{svg}
\usepackage{cite}
\usepackage{subcaption}

\usepackage{booktabs} 
\usepackage{placeins}
\usepackage{color}

\usepackage{amsmath}
\usepackage{algorithm}
\usepackage{algpseudocode}
\usepackage{array,booktabs}
\usepackage{flushend}

\usepackage{multirow}

\usepackage{tabularx}
\usepackage[export]{adjustbox}
\usepackage{rotating}

\makeatletter
\let\NAT@parse\undefined
\makeatother
\usepackage{hyperref}
\usepackage{xcolor}
\usepackage{wrapfig}
\usepackage{orcidlink}
\usepackage{cleveref}
\usepackage{siunitx}

\newcommand{\PreserveBackslash}[1]{\let\temp=\\#1\let\\=\temp}
\newcolumntype{C}[1]{>{\PreserveBackslash\centering}p{#1}}
\newcolumntype{R}[1]{>{\PreserveBackslash\raggedleft}p{#1}}
\newcolumntype{L}[1]{>{\PreserveBackslash\raggedright}p{#1}}

\definecolor{OrangeCR}{HTML}{f1b514}
\definecolor{GreenCR}{HTML}{008000}
\definecolor{GreenForce}{HTML}{00b050}
\definecolor{red}{HTML}{ff0000}
\definecolor{TUMBlue}{HTML}{0065bd}
\definecolor{TUMOrange}{HTML}{E37222}
\definecolor{TUMGreen}{HTML}{9FBA36}
\definecolor{lightOrange}{HTML}{FFC000}
\definecolor{lightGreen}{HTML}{92D050}
\definecolor{GrayArrow}{HTML}{838383}
\definecolor{Grey}{HTML}{808080}

\definecolor{policy-road}{HTML}{c04f15}
\definecolor{policy-sidewalk}{HTML}{fbe3d6}
\definecolor{policy-crosswalk}{HTML}{f2aa84}
\definecolor{policy-goal}{HTML}{4ea72e}
\usetikzlibrary{arrows, arrows.meta, external, matrix}

\newcommand{\boxCR}{
  \coloredbox{OrangeCR}
}

\newcommand{\boxSidewalk}{
  \coloredbox{lightGreen}
}

\newcommand{\boxCrosswalk}{
  \coloredbox{lightOrange}
}

\newcommand{\coloredbox}[1]{
  \tikz{
    \draw[fill=#1] (0,0.0) rectangle (0.4,0.15);
  }
}

\newcommand{\coloredsquare}[1]{
  \tikz{
    \draw[fill=#1] (0,0.0) rectangle (0.3,0.3);
  }
}

\newcommand{\circlePedestrianGoal}[1]{
  \tikz{
    \draw[fill=#1] (0,0) circle (1.7pt);
  }
}

\newcommand{\arrowCR}{
  \tikz{
    \draw[{Circle[GreenCR,length=4pt]}-{triangle 45 [GreenCR,length=4pt]}, GreenCR] (0,0) -- (0.5,0);
    \draw[line width=2pt, GreenCR] (0.05,0) -- (0.3,0);
  }
}

\newcommand{\boxPedestrian}{
  \tikz{
    \draw[fill=TUMBlue] (0,0.0) rectangle (0.2,0.1);
  }
}

\newcommand{\boxPedestriancolor}[1]{
  \tikz{
    \draw[fill=#1] (0,0.0) rectangle (0.2,0.1);
  }
}

\newcommand{\forceArrow}[1]{
  \tikz{
    \draw[-{Stealth[length=5pt,width=6pt]}, line width=1.5pt, #1] (0,0) -- (0.5,0);
  }
}

\newcommand{\dashedGrayArrow}{
  \tikz{
    \draw[-{Triangle[length=5pt,width=7pt]}, line width=3pt, GrayArrow, dash pattern=on 3pt off 0.8pt] (0,0) -- (0.5,0);
  }
}

\newcommand{\UncertainPrediction}{
  \tikz{
    \fill[yellow,opacity=0.5] (0,0) circle (0.18);
    \fill[orange,opacity=0.5] (0,0) circle (0.1);
    \fill[red,opacity=0.5] (0,0) circle (0.04);
  }
}

\begin{document}


\title{Pedestrian-Aware Motion Planning for Autonomous Driving in Complex Urban Scenarios}

\author{Korbinian Moller$^{\orcidlink{0000-0001-7120-0796}1}$, Truls Nyberg$^{\orcidlink{0000-0002-2069-6581}2}$, Jana Tumova$^{\orcidlink{0000-0003-4173-2593}2}$, Johannes Betz$^{\orcidlink{0000-0001-9197-2849}1}$\vspace{-1.5\baselineskip}
\thanks{$^1$ The authors are with the Professorship of Autonomous Vehicle Systems (AVS), TUM School of Engineering and Design, Technical University of Munich, 85748 Garching, Germany; Munich Institute of Robotics and Machine Intelligence (MIRMI).}
\thanks{$^2$ The authors are with the Division of Robotics, Perception and Learning (RPL), School of Electrical Engineering and Computer Science, KTH Royal Institute of Technology, 10044 Stockholm, Sweden.}
}

\markboth{}
{Moller \MakeLowercase{\textit{et al.}}: Motion Planning with Uncertain Pedestrian Behavior}

\IEEEpubid{}

\maketitle


\begin{abstract}
Motion planning in uncertain environments like complex urban areas is a key challenge for autonomous vehicles (AVs). The aim of our research is to investigate how AVs can navigate crowded, unpredictable scenarios with multiple pedestrians while maintaining a safe and efficient vehicle behavior. So far, most research has concentrated on static or deterministic traffic participant behavior. This paper introduces a novel algorithm for motion planning in crowded spaces by combining social force principles for simulating realistic pedestrian behavior with a risk-aware motion planner.
We evaluate this new algorithm in a 2D simulation environment to rigorously assess AV-pedestrian interactions, demonstrating that our algorithm enables safe, efficient, and adaptive motion planning, particularly in highly crowded urban environments—a first in achieving this level of performance. This study has not taken into consideration real-time constraints and has been shown only in simulation so far. Further studies are needed to investigate the novel algorithm in a complete software stack for AVs on real cars to investigate the entire perception, planning and control pipeline in crowded scenarios. We release the code developed in this research as an open-source resource for further studies and development. It can be accessed at the following link: \url{https://github.com/TUM-AVS/PedestrianAwareMotionPlanning}

\end{abstract}




\section{Introduction}
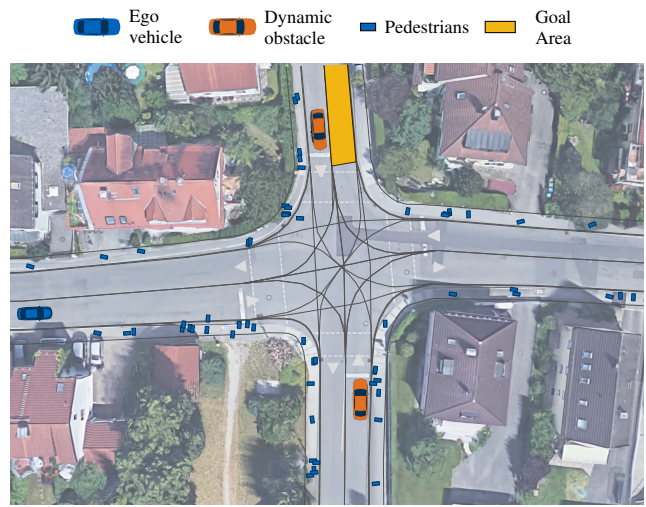
\begin{figure}[!t]
    \centering
    \begin{tikzpicture}[font=\scriptsize]
        \node[inner sep=0pt] at (0.9,0) {\includegraphics[height=2.5mm]{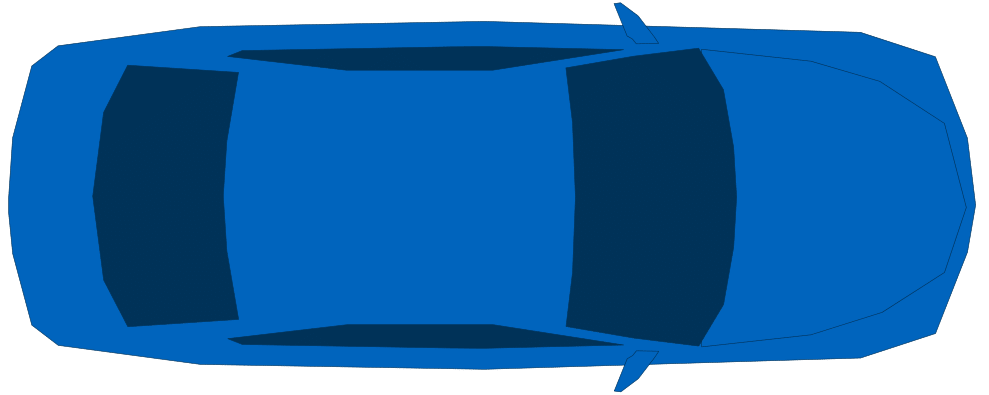}};
        \node[align=left, anchor=west] at (1.2,0) {Ego \\ vehicle};
    
        \node[inner sep=0pt] at (2.7,0) {\includegraphics[height=2.5mm]{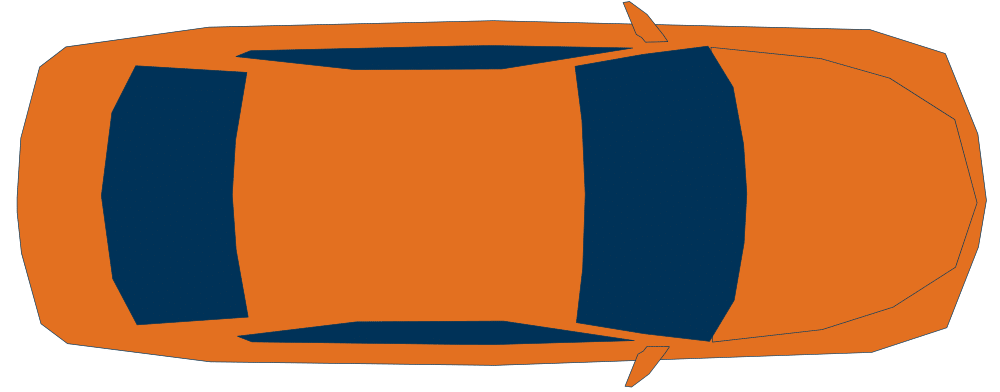}};
        \node[align=left, anchor=west] at (3.0,0) {Dynamic \\ obstacle};

        \node[inner sep=0pt] at (4.5,0) {\boxPedestrian};
        \node[align=left, anchor=west] at (4.6,0) {Pedestrians};

        \node[inner sep=0pt] at (6.3,0) {\boxCR};
        \node[align=left, anchor=west] at (6.6,0) {Goal \\ Area};
    \end{tikzpicture}
    \includegraphics[width=0.95\linewidth, trim={2cm 3cm 8cm 0cm},clip]{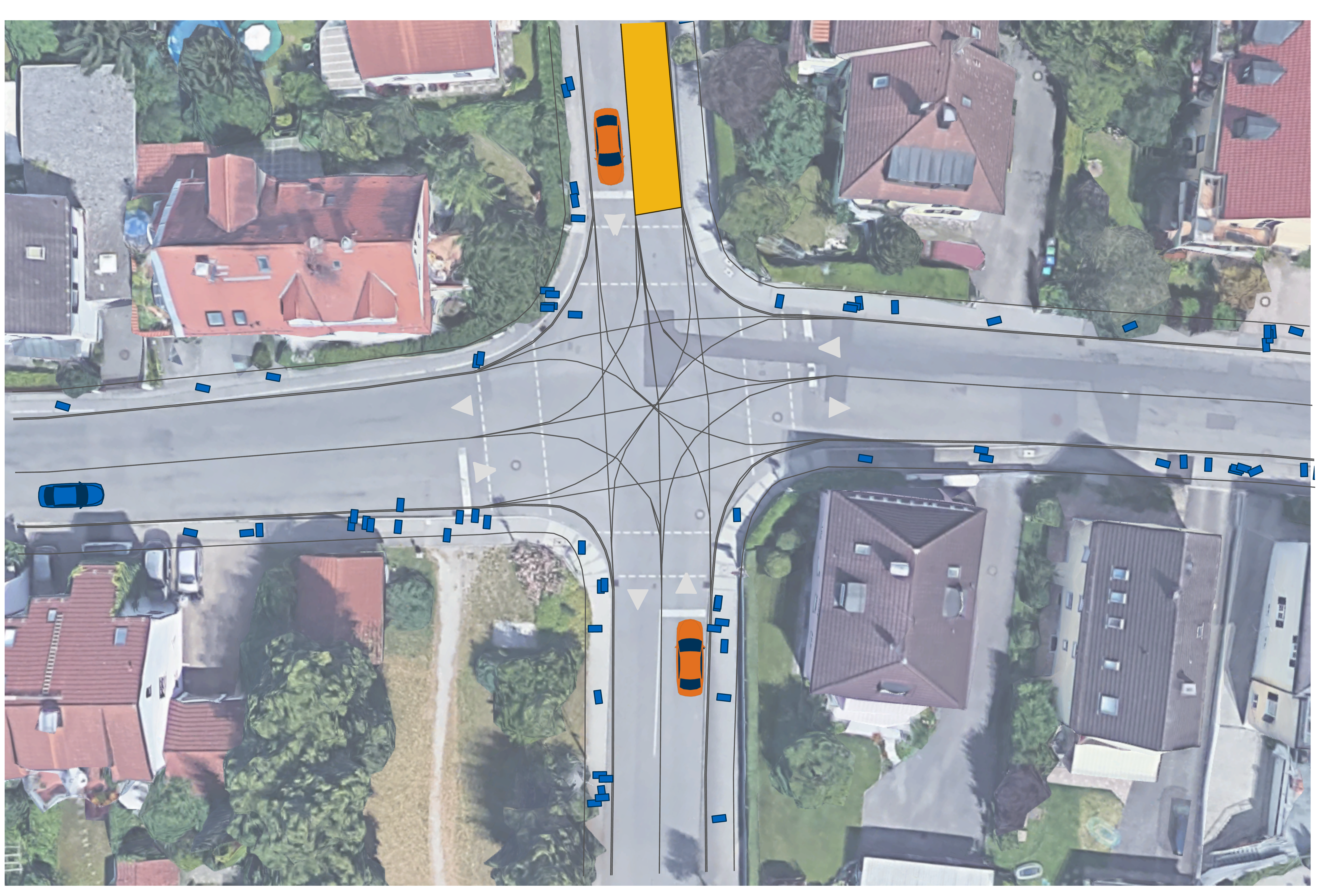}
    \caption{Example of a complex urban intersection illustrating the challenge for an AV to navigate safely and efficiently around a high density of pedestrians and human-driven vehicles}
    \label{fig:urbanenvironment}
\end{figure}

In a world increasingly shaped by technology, autonomous vehicles (AVs) represent a transformative step in mobility, with the potential to revolutionize traffic systems and transportation habits~\cite{Bobisse2019}. The underlying autonomy algorithms \cite{pendleton2017perception} must be designed to respond to their environment in real time to create safe and efficient vehicle behavior.
Despite the promises of autonomous driving, practical experiences, and various collision reports have highlighted significant challenges that this technology must overcome in the real world~\cite{Pokorny2022}.
As the operational design domain (ODD) \cite{Taxonomy2024} of AVs expands beyond highways and controlled environments to include complex urban areas (\Cref{fig:urbanenvironment}), the challenge of dealing with uncertain and unpredictable behavior of traffic participants, e.g., pedestrians, becomes more profound~\cite{Huang.2023, rasouli_autonomous_2020}. 

This paper addresses the challenges of highly reactive motion planning in urban environments by developing and evaluating a risk-aware motion planner for pedestrian-rich environments. 
By evaluating potential harm and collision probabilities, the planner ensures safe navigation while preventing overly cautious behavior, thereby mitigating the robot freezing problem.

Further, to the best of our knowledge, existing simulation environments like CARLA \cite{Dosovitskiy17} or CommonRoad \cite{althoff_commonroad_2017} lack the capability to realistically model pedestrian behavior, limiting their applicability for the development of advanced motion planning algorithms. To address this need, we introduce a dedicated pedestrian simulation model that operates on top of 2D motion planning environments. This model leverages a social force approach to simulate interactive and dynamic pedestrian behavior, enabling realistic interactions with AVs and providing a crucial foundation for the development and evaluation of advanced motion planning algorithms.

\IEEEpubidadjcol 

Through extensive simulations, we validate the effectiveness of our pedestrian simulation model and the pedestrian-aware motion planner, demonstrating improved AV safety and performance in complex urban settings.
In conclusion, this work provides four key contributions:

\begin{itemize}
    \item A risk-aware motion planning algorithm that evaluates collision probability and potential harm, enabling safe trajectory planning in pedestrian-rich environments while mitigating the freezing robot problem.
    \item A novel pedestrian simulation model based on social force dynamics, which provides an adaptable environment for developing and testing AV planning algorithms in realistic urban scenarios.
    \item A comprehensive evaluation of the proposed risk-aware motion planner in pedestrian-rich scenarios, showcasing its ability to manage unpredictable pedestrian behavior effectively. 
    \item We provide open-source access to the motion planner and pedestrian simulation model.
\end{itemize}

\section{Related Work}

\subsection{Behavior Modeling and Prediction}
\label{subsec:PedestrianBehavior}
With advancements in autonomous systems, regulations are emerging for vehicles with increasing levels of automation, such as the UN Regulation 157 for Automated Lane Keeping Systems (ALKS)~\cite{UN_Regulation_157} and the European Regulation 1426/2022 governing Automated Driving Systems (ADS)~\cite{ciuffo_2024}. These regulations define safety requirements for interactions between vehicles and other road users. To support these requirements, a holistic scenario understanding and motion prediction \cite{karle2022scenario} like behavior models of traffic participants are needed.

However, behavior models for vehicle-to-pedestrian interactions remain underdeveloped and are not yet standardized. Although new cars sold in Europe are required to have Advanced Emergency Braking Systems (AEBS) capable of detecting and braking for vehicles, bicyclists, and pedestrians~\cite{EU_Regulation_2019_2144}, these systems are designed to assist human drivers rather than replace them entirely. While AEBS can improve safety, they rely on simple behavior models and are insufficient to prevent all accidents, particularly in complex urban environments where pedestrian behavior is unpredictable~\cite{haus_estimated_2019, CICCHINO2022106686}.

Driverless AVs face unique challenges when interacting with pedestrians, whose behavior can be unpredictable, culturally variable~\cite{sprenger_cross-cultural_2023}, and challenging to model~\cite{camara_pedestrian_2021-1, camara_pedestrian_2021-2}. Numerous surveys have explored pedestrian prediction methods for AVs, ranging from physics-based approaches to deep learning models~\cite{rudenko_human_2020, gulzar_survey_2021, sharma_pedestrian_2022, zhang_pedestrian_2023}. While more sophisticated techniques, such as Social-LSTM and Social-GAN, excel in predicting trajectories in controlled environments, they often underperform in real-world scenarios with dense crowds or rapid behavior changes~\cite{korbmacher_review_2022, korbmacher_deep_2024}. Recent approaches have focused on integrating contextual and environmental information into prediction models to better capture pedestrian behavior under complex urban conditions \cite{kalatian2022context, azarmi2023local, fang2024behavioral}. 

\subsection{Motion Planning for AV}
\label{subsec:MotionPlanning}
The motion planning process in autonomous driving software can be defined as determining a feasible path and the corresponding time-dependent actions for a vehicle to reach its destination safely and efficiently. In the literature, motion planning algorithms are broadly categorized into optimization-, graph-, sampling-, and machine learning-based approaches~\cite{Paden2016,Teng2023,Zhou2022,Gonzlez2015,Dong2023,Tampuu2022}. Here, we focus specifically on sampling-based motion planners, as they provide an effective balance between computational efficiency and flexibility in dynamic environments ~\cite{Paden2016}. In sampling-based planners, trajectories are generated by connecting sampled states to the vehicle's current state and evaluated using cost functions, which may include factors like travel time, comfort, or collision probability for safe navigation in uncertain environments~\cite{Frenetix, nyberg2021risk, trauth2023toward, Moller2024, Piazza2024}.


Nevertheless, motion planning in shared spaces demands robust strategies to model human behavior and navigate dynamic, crowded environments safely~\cite{Monderman2006}.

In \cite{Liu2015}, the authors propose a hierarchical planning system that uses LIDAR data for traversability analysis in pedestrian-rich areas. While effective for fundamental obstacle avoidance, the approach treats pedestrians as static obstacles, failing to predict their future motion or intentions. Similarly, Morales et al.~\cite{Morales2016} combine global and local planning with cost maps to handle static and dynamic obstacles.

Yang et al.~\cite{Yang2023} propose a framework that integrates pedestrian motion prediction using LSTM networks with Frenet-based trajectory planning. However, while the approach effectively models sequential pedestrian motion, it focuses primarily on ensuring safety through safety margins without employing more sophisticated methods to account for broader interactions or risk-awareness in dense environments.

Partially Observable Markov Decision Processes (POMDPs) provide a probabilistic framework to manage uncertainty in pedestrian behavior. \cite{Bai2015, Luo2018} use POMDPs to estimate pedestrian intentions within a hierarchical planning system. However, they simplify pedestrian motion by assuming straight-line trajectories. 

Li et al.~\cite{Li2024} address occlusion challenges with a Stochastic Model Predictive Control (SMPC) framework. By incorporating phantom pedestrian models, the system quantifies uncertainties in occluded areas, enabling safer navigation. Similarly,\cite{Zhu2022} explores hybrid models combining reinforcement learning with rule-based constraints to handle distracted pedestrian interactions at unsignalized crosswalks. While both methods achieve safety improvements, their reliance on simplified pedestrian motion assumptions limits their effectiveness in densely populated environments.

In \cite{Li2020}, the authors propose a framework to combine reinforcement learning with multimodal trajectory prediction using Social GAN. By extending the action space and incorporating kinematic constraints, the approach ensures natural and human-like AV trajectories. However, it heavily penalizes potential collisions, emphasizing reactive safety measures over long-term pedestrian-vehicle interactions.

While many of the existing methods focus on ensuring safety through collision probability-based approaches, these frameworks are insufficient for pedestrian-dense environments. In this context, harm- and risk-aware motion planning approaches offer promising solutions by explicitly incorporating the potential risk into the planning process~\cite{geisslinger2021autonomous}, ultimately enabling the generation of safer trajectories.


\subsection{Simulation and Validation}
\label{subsec:Simulation}
Given the challenges of integrating pedestrian behavior into AV motion planning, simulation tools play a crucial role in testing and validating the effectiveness of planners.
However, existing platforms often struggle to model reactive pedestrian behaviors effectively, which limits their ability to provide realistic evaluations.

Open datasets such as nuScenes~\cite{nuScenes} provide extensive real-world driving logs but lack the dynamic feedback needed for thorough planner assessment. 
Open-loop evaluations like Average Displacement Error (ADE) and Final Displacement Error (FDE) focus on ego-forecasting accuracy without considering interactions with other agents. 
In~\cite{Dauner2023CORL}, the authors critique the overreliance on open-loop metrics, highlighting how nuPlan~\cite{caesar_nuplan_2022} planners perform well in open-loop tests but struggle when real-time feedback is introduced.

Simulation platforms like nuPlan~\cite{caesar_nuplan_2022} and CommonRoad~\cite{althoff_commonroad_2017} are increasingly popular for validating AV planners, offering both open- and closed-loop capabilities. nuPlan, which uses the Intelligent Driver Model (IDM)~\cite{idm} for reactive vehicle behaviors, has limited pedestrian modeling, with non-reactive, predetermined paths. While~\cite{hallgarten_can_2024} introduced jaywalking pedestrians for more challenging test cases, these pedestrians still do not respond in real time to the AV’s actions. Similarly,~\cite{chitta_sledge_2024} proposes generating longer-duration scenarios, but pedestrian behavior remains static.

A promising alternative for pedestrian modeling is the Social Force Model~\cite{helbing_social_1995}, widely used in crowd dynamics. Notable variants include work on signalized crosswalks~\cite{zeng_modified_2014}, general waiting behavior~\cite{johansson_waiting_2015}, and deep learning-based adaptations~\cite{kreiss_deep_2021}, which improve the classical model by considering more complex pedestrian interactions. Despite these improvements, fully integrating the Social Force Model into urban traffic scenarios remains challenging.

SUMO~\cite{SUMO2018} is actively developing the social force model for pedestrian simulation~\cite{chraibi_jupedsim_2024}, and efforts like~\cite{klischat_coupling2019} show promise by coupling SUMO with CommonRoad for more realistic closed-loop simulations. However, these developments are still not fully available for testing. To address this gap, we integrate the Social Force Model into CommonRoad~\cite{althoff_commonroad_2017}, enabling realistic pedestrian-vehicle interactions and providing a robust testbed for evaluating AV planners in complex urban conditions.


\section{Problem Formulation}

\subsection{Motion Planning Problem}
\begin{figure}[!htpb]
    \centering
    \begin{tikzpicture}[font=\scriptsize]
        \node[inner sep=0pt] at (0.7,0) {\includegraphics[height=2.5mm]{Images/icons/ego.png}};
        \node[align=left, anchor=west] at (1.0,0) {Ego \\ vehicle};
    
        \node[inner sep=0pt] at (2.5,0) {\includegraphics[height=2.5mm]{Images/icons/dyn_obstacle.png}};
        \node[align=left, anchor=west] at (2.8,0) {Dynamic \\ obstacle};

        \node[inner sep=0pt] at (4.3,0) {\boxPedestrian};
        \node[align=left, anchor=west] at (4.4,0) {Pedestrians};

        \node[inner sep=0pt] at (6.1,0) {\arrowCR};
        \node[align=left, anchor=west] at (6.35,0) {Start $x_0$};

        \node[inner sep=0pt] at (7.7,0) {\boxCR};
        \node[align=left, anchor=west] at (7.9,0) {Goal $\mathcal{G}$};

        \node[anchor=west, inner sep=0pt] at (0.35,-1.25) {\includegraphics[width=0.95\linewidth, trim={15cm 0cm 14cm 0cm},clip]{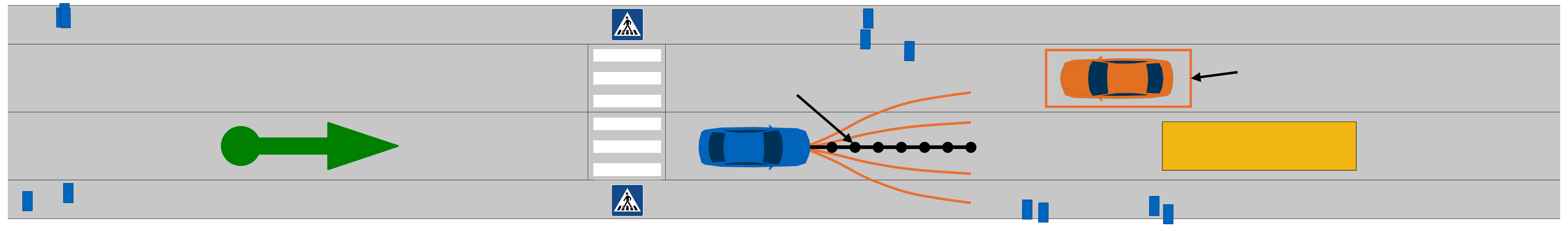}};
        
        \node[align=left, anchor=west] at (7.7,-0.95) {$\mathcal{C}_\text{coll}$};   

        \node[align=left, anchor=west] at (3.7,-0.98) {$x(t_2), y(t_2)$};
        
    \end{tikzpicture}
    \caption{Illustration of the motion planning problem. The ego vehicle plans its trajectory while considering dynamic obstacles and pedestrians. 
    A sample collision constraint $\mathcal{C}_\text{coll}$, along with computed points of the selected trajectory and other non-selected trajectory candidates (orange), is shown.}
    \label{fig:motionplanningproblem}
\end{figure}

The objective of this work is to address the challenge of motion planning in pedestrian-rich environments, as shown in \Cref{fig:motionplanningproblem}. 
This involves determining a safe and efficient trajectory $\tau(t)$ from an initial vehicle state $x_0$ to a goal region $\mathcal{G}$, all while respecting various constraints $\mathcal{C}$, such as collision avoidance $\mathcal{C}_\text{coll}$ and others specific to the task and environment. The trajectory, defined by spatial coordinates $x(t), y(t)$ and a velocity profile $v(t)$, acts as a reference for the low-level vehicle control. A cost function $J(\tau)$ evaluates each trajectory based on criteria such as efficiency, comfort, and overall travel time, ensuring an optimal solution for the given scenario.


\subsection{Pedestrian Simulation Problem}
The pedestrian simulation problem addresses the challenge of replicating realistic pedestrian motion in structured environments. Pedestrians move from their starting locations $s_i$ to their target destinations $g_i$ within urban settings characterized by sidewalks (SW) and crosswalks (CW). SWs provide primary pathways, while CWs are used to safely traverse streets. Pedestrians naturally seek to avoid collisions by respecting personal space and dynamically responding to the presence of others~\cite{Corbetta2018}. 

\begin{figure}[!ht]
    \centering
    \begin{tikzpicture}[font=\scriptsize]
        \node[inner sep=0pt] at (1.0,0) {\includegraphics[height=2.5mm]{Images/icons/ego.png}};
        \node[align=left, anchor=west] at (1.5,0) {Ego \\ vehicle};
    
        \node[inner sep=0pt] at (2.8,0) {\includegraphics[height=2.5mm]{Images/icons/dyn_obstacle.png}};
        \node[align=left, anchor=west] at (3.1,0) {Dynamic \\ obstacle};

        \node[inner sep=0pt] at (4.5,0.15) {\boxPedestriancolor{black}};
        \node[inner sep=0pt] at (4.5,0) {\boxPedestriancolor{TUMBlue}};
        \node[inner sep=0pt] at (4.5,-0.15) {\boxPedestriancolor{TUMOrange}};
        \node[align=left, anchor=west] at (4.6,0) {$s_i$};

        \node[inner sep=0pt] at (5.3,0.15) {\circlePedestrianGoal{black}};
        \node[inner sep=0pt] at (5.3,0) {\circlePedestrianGoal{TUMBlue}};
        \node[inner sep=0pt] at (5.3,-0.15) {\circlePedestrianGoal{TUMOrange}};
        \node[align=left, anchor=west] at (5.4,0) {$g_{1-3}$};

        \node[inner sep=0pt] at (6.5,0) {\boxSidewalk};
        \node[align=left, anchor=west] at (6.7,0) {SW};

        \node[inner sep=0pt] at (7.7,0) {\boxCrosswalk};
        \node[align=left, anchor=west] at (7.9,0) {CW};

        \node[anchor=west, inner sep=0pt] at (0.35,-1.25) {\includegraphics[width=0.95\linewidth, trim={15cm 0cm 14cm 0cm},clip]{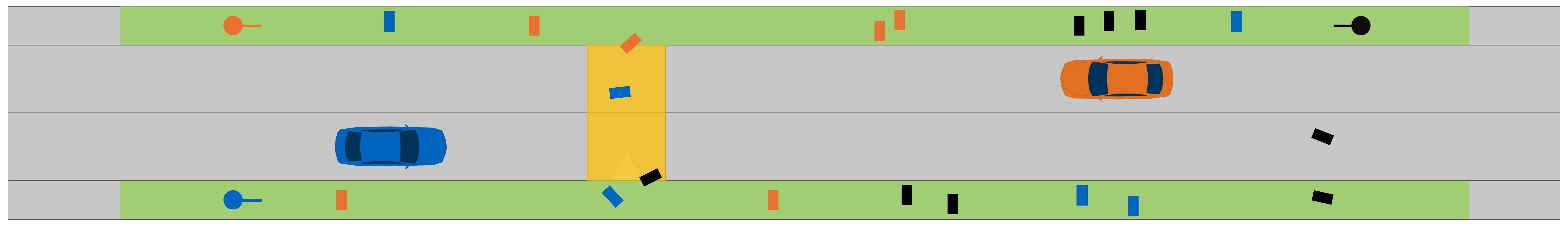}};

        \node[align=left, anchor=west] at (8.0,-0.62) {$g_1$};

        \node[align=left, anchor=west] at (0.72,-0.62) {$g_3$};

        \node[align=left, anchor=west] at (0.72,-1.88) {$g_2$};
    
    \end{tikzpicture}
    \caption{Illustration of the pedestrian simulation problem at the initial time step ($t=0$). Pedestrians are shown with their respective starting positions ($s_i$) and are grouped based on their shared target locations ($g_{1-3}$), with each group and respective target represented by a unique color. While pedestrians within the same group share a common goal, they act as independent agents. The illustration highlights key areas such as SWs and CWs and showcases behaviors like illegal road crossings, which occur when pedestrians are far from CWs, and their destination lies across the street.}
    \label{fig:pedestriansimulationproblem}
\end{figure}

Interactions are governed by social dynamics and physical constraints, such as avoiding obstacles and vehicles. The latter are perceived as high-risk elements, leading pedestrians to maintain larger safety distances. These behaviors ensure a balance between individual goals (e.g., reaching $g_i$) and environmental factors (e.g., traffic regulations). Pedestrian movement is defined by a trajectory $\tau_\mathrm{ped}(t)$ consisting of position coordinates $x_\mathrm{ped}(t)$ and $y_\mathrm{ped}(t)$ and a velocity profile $v_\mathrm{ped}(t)$. 
An example scenario is depicted in \Cref{fig:pedestriansimulationproblem}.

\section{Methodology}

This chapter outlines the methodology and details of the implementation of the pedestrian-aware motion planner and the proposed pedestrian simulator.
The framework and its step-by-step workflow are introduced in \Cref{subsec:Framework}, followed by a detailed breakdown of each module. \Cref{subsec:PedestrianSimulator} comprehensively explains the pedestrian simulation, covering the key calculations and behavioral models. Subsequently, \Cref{subsec:Pedestrian-awareMotionPlanning} describes the pedestrian-aware motion planning approach.

\subsection{Framework}
\label{subsec:Framework}

\begin{figure*}[!ht]
    \centering
    \input{Images/Framework.tex}
    \caption{Overview of the simulation framework. The iterative process integrates a pedestrian simulator and a motion planner to evaluate safe and efficient vehicle trajectories in complex urban environments.}
    \label{fig:framework}
\end{figure*}
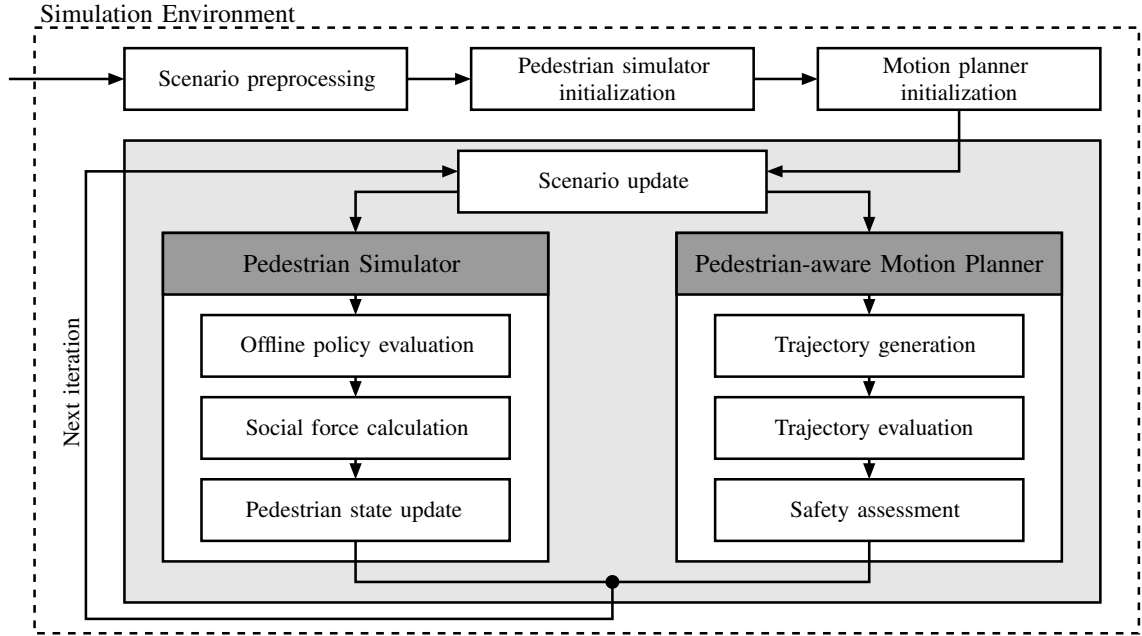

The developed pedestrian simulation model aims to simulate realistic pedestrian behavior in urban environments to support the evaluation and development of pedestrian-aware motion planners.

As shown in \Cref{fig:framework}, the framework begins with scenario preprocessing and initializes its two main components: the pedestrian simulator and the motion planner. Once initialized, these components engage in an iterative, synchronous simulation loop within a continuously evolving scenario.

The pedestrian simulator leverages a Social Force Model to generate realistic pedestrian movements, accounting for both individual behaviors and interactions with the environment and other agents. Each iteration updates pedestrian states in response to vehicles and other pedestrians.

The pedestrian-aware motion planner uses sampling-based techniques to generate and evaluate multiple trajectory candidates. The selected trajectory updates the vehicle’s state, including its position, velocity, and heading.

After both the pedestrian simulator and the motion planner have completed their computations for each time step, collision checks are performed at the simulation level to ensure no collisions have occurred. Each iteration represents a short time horizon, typically $\Delta t = \SI{0.1}{\second}$, providing a detailed evaluation of motion planning algorithms in dynamic, realistic environments.

\subsection{Pedestrian Simulation}
\label{subsec:PedestrianSimulator}
Our pedestrian simulation model is integrated into a 2D simulation environment, enhancing it with pedestrian-specific elements. Where SWs are missing, they are added along existing roads to align with the street layout. Similarly, CWs are manually introduced at appropriate locations within the scenario. All added elements are represented as 2D polygons. Pedestrian clusters are then randomly generated along the SWs, with their spacing determined by an exponential distribution based on a user-defined average distance between clusters. The number of pedestrians in each cluster follows a geometric distribution, with a defined average of pedestrians per cluster. Each pedestrian is assigned an initial orientation toward their goal and a parameterized desired velocity. Their positions are normally distributed around the cluster center. Pedestrians whose positions fall outside the SW boundary are not spawned.

The motion of pedestrians is based on the social force model. The implementation is inspired by previous implementations\cite{kreiss_deep_2021, gao_yuxiang-gaopysocialforce_2024, noauthor_srl-freiburgpedsim_ros_2024, chraibi_jupedsim_2024}, but modified to work in a structured traffic environment. 
The social force model describes human motion using attractive and repulsive forces. An attractive force is used to steer the pedestrian towards its goal, and between each pedestrian and obstacle, a repulsive social force is modeled for collision avoidance. In this work, we also add a repulsive force from vehicles. 

In the social force model, the attractive force is often just computed by using the directional vector from a pedestrian's current position towards its goal. 
This may work well in open areas, where pedestrians want to follow the shortest path towards their goal. 
However, in a traffic environment with SWs and crossings, more aspects must be considered to find pedestrians' preferred and desired walking directions. 
A naïve modification to the social force model would be to add attractive forces to dedicated walking areas and repulsive forces to lanes. 
This may work in certain situations, but it will generally lead to a potential field with local minima where pedestrians get stuck.

\begin{figure}[H]
    \centering
    \begin{tikzpicture}[font=\scriptsize]
    


        \node[inner sep=0pt] at (1.1,0) {\dashedGrayArrow};
        \node[align=left, anchor=west] at (1.4,0) {Vehicle \\ prediction};

        \node[inner sep=0pt] at (3.1,0) {\forceArrow{GreenForce}};
        \node[align=left, anchor=west] at (3.4,0) {Attractive \\ forces};

        \node[inner sep=0pt] at (5.2,0) {\forceArrow{red}};
        \node[align=left, anchor=west] at (5.5,0) {Social \\ forces};

        \node[inner sep=0pt] at (7.0,0) {\forceArrow{TUMBlue}};
        \node[align=left, anchor=west] at (7.3,0) {Vehicle \\ forces};

        \node[anchor=west, inner sep=0pt] at (0.35,-1.6) {\includegraphics[width=0.95\linewidth, trim={0.05cm 0.3cm 0.05cm 0cm},clip]{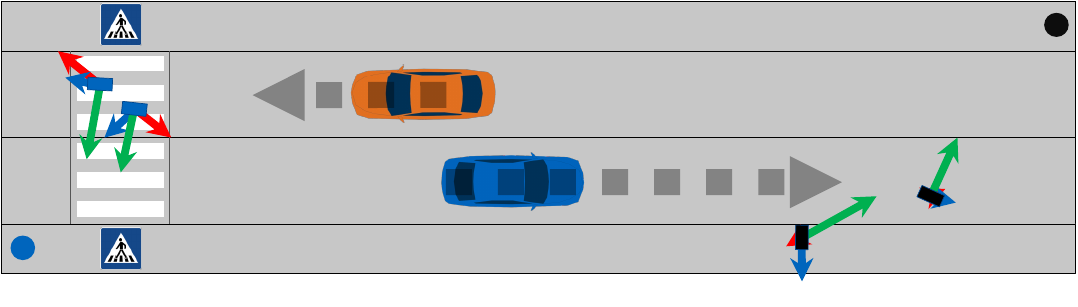}};
    
    \end{tikzpicture}
    \caption{Illustration of the forces influencing pedestrian motion in the scenario. The blue pedestrians are navigating toward their goal in the bottom-left corner, while the black pedestrians aim for their goal in the top-right corner, requiring them to cross the street. The gray dashed lines represent the vehicle predictions, which influence pedestrian behavior.}
    \label{fig:pedestrianforce}
\end{figure}
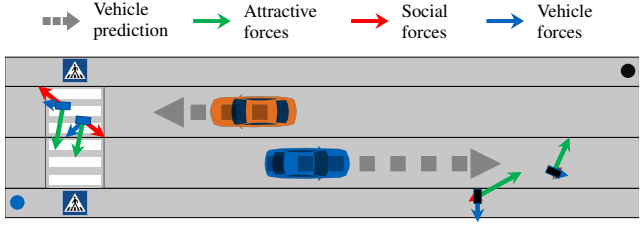

Figure~\ref{fig:pedestrianforce} illustrates the forces in a scenario where pedestrians are trying to cross a street. Colored arrows depict the attractive and repulsive forces acting on the pedestrians. The two pedestrians at the CW exert opposing social forces on each other, but because the one behind is outside the field of view of the one ahead, the force from the pedestrian behind is scaled down. The slowly approaching orange vehicle exerts a weak force on the pedestrians at the CW since it is not predicted to reach far. In contrast, the faster-moving blue vehicle exerts a stronger force, discouraging one pedestrian from jaywalking. The other pedestrian is far enough away to not be affected.

To tackle the issue of local minima, pedestrians must plan with a longer horizon. They must be able to negotiate whether to cross a lane immediately or take a detour to a dedicated crossing. 
This planning problem can be solved in many ways, but solving it in real time can be challenging, especially when the number of pedestrians grows. 
In this work, we solve a simple planning problem offline to obtain a policy for a given set of possible goal positions. 
The policy gives a desired direction, given any current position. 
The desired direction can then be used online to compute a force together with the other social forces.

To compute the offline policies, we use value iteration with a discretized set of actions over a discretized grid of positional states.
The actions are the direction vectors to the $n$ closest cell centers with unique angles, each with a cost equal to the length of the resulting movement. 
A state cost is also introduced to avoid lanes and prioritize SWs and CWs. This is done by rasterizing the lanelet polygons into a grid and assigning decreasing costs for roads, crossings, and SWs, respectively (e.g., cost of 50, 20, and 10).

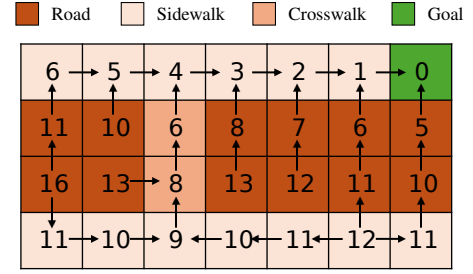
\begin{figure}[!ht]
    \centering
    \begin{tikzpicture}[font=\scriptsize]
        \node[inner sep=0pt] at (1.8,0) {\coloredsquare{policy-road}};
        \node[align=left, anchor=west] at (2.0,0) {Road};
    
        \node[inner sep=0pt] at (3.2,0) {\coloredsquare{policy-sidewalk}};
        \node[align=left, anchor=west] at (3.4,0) {Sidewalk};

        \node[inner sep=0pt] at (4.95,0) {\coloredsquare{policy-crosswalk}};
        \node[align=left, anchor=west] at (5.15,0) {Crosswalk};

        \node[inner sep=0pt] at (6.8,0) {\coloredsquare{policy-goal}};
        \node[align=left, anchor=west] at (7.0,0) {Goal};

        \node[anchor=west, inner sep=0pt] at (1.7,-1.9) {\includegraphics[width=0.65\linewidth]{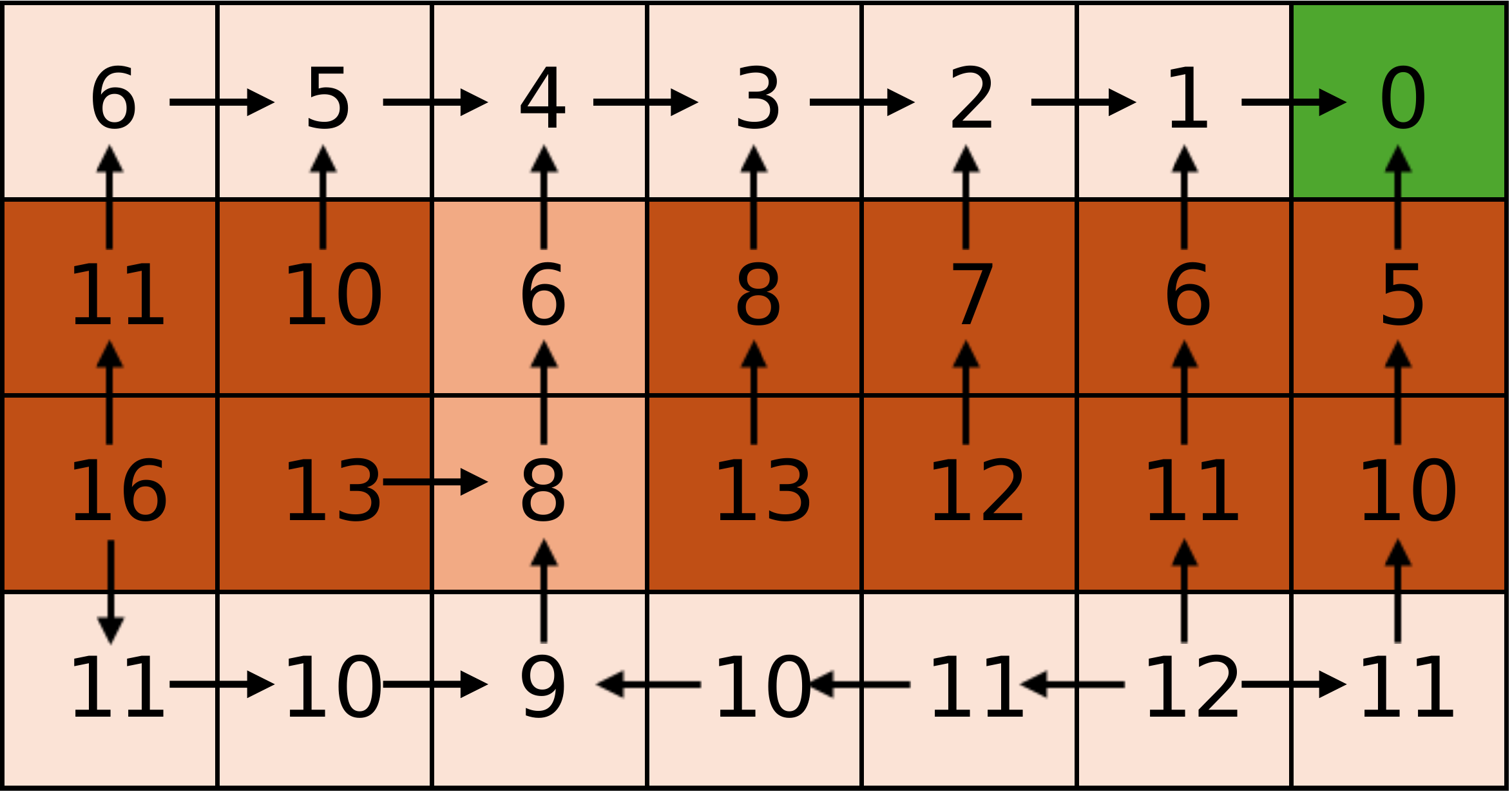}};
    
    \end{tikzpicture}
    \caption{Illustration of a simple policy for pedestrians moving toward the top-right sidewalk. Cell colors indicate state costs, and numbers show the total cost-to-go from each state.}
    \label{fig:policy}
\end{figure}

For simplicity, the transitions are assumed to be deterministic and made at a constant speed (although, in practice, pedestrians will also be influenced by social forces). 
Transitions to non-neighboring cells are allowed to increase the angular granularity in the action space. 
The state cost is scaled with the transition length and assumed constant during the action to account for passing through multiple states.
By running the classical value iteration algorithm, a policy can be obtained with the best action from each discretized state~\cite{thrun2005probabilistic}. A simple example is illustrated in Figure~\ref{fig:policy}, where the pedestrian policy for reaching the top right corner is visualized with the resulting cost-to-go values from each discretized position in the grid.

With a pedestrian's desired direction from the policy, $e_\alpha^\pi$, the attractive force, $F_\alpha^0$, can be calculated as proposed in the original social force model as
\begin{equation}
    F_\alpha^0 = \frac{ e_\alpha^\pi v_\alpha^0 - v_\alpha}{\tau},
\end{equation}
where $v_\alpha^0$ is a desired target velocity, $v_\alpha$ the current velocity, and $\tau$ is a relaxation time. 

The social force between two pedestrians $\alpha$ and $\beta$ is calculated from a repulsive exponentially decreasing potential $V_{\alpha\beta}(b) = V_{\alpha\beta}^0\exp\left(-b/\sigma_\beta\right)$ with
\begin{equation}
    b(r_{\alpha\beta}) = 0.5 \sqrt{(||r_{\alpha\beta}|| + ||r_{\alpha\beta} - v_\beta s_\beta ||)^2 + ||v_\beta s_\beta||^2}
\end{equation}
where $r_{\alpha\beta}$ is the vector between the pedestrian positions, i.e., $p_\beta(t) - p_\alpha(t)$, $v_\beta$ is the velocity vector of pedestrian $\beta$, and $s_\beta$ is the step width of pedestrian $\beta$.
The force is then given as
\begin{equation}
    f_{\alpha\beta}(r_{\alpha\beta}) = -\nabla_{r_{\alpha\beta}} V_{\alpha\beta}[b(r_{\alpha\beta})]
\end{equation}
which we compute with a finite difference approximation.

For the repulsive force on pedestrians from a vehicle $\gamma$, we use a simpler exponential potential
\begin{equation}
    V_{\alpha\gamma}(r_{\alpha\gamma}) = - V_\gamma^0 \exp \left(\frac{||r_{\alpha \gamma'}||}{\sigma_\gamma}\right),
\end{equation}
with parameters $V_\gamma^0$ and $\sigma_\gamma$, which gives an analytically tractable gradient. 
To account for vehicles' higher velocity and repel pedestrians also in front of vehicles, we compute the potential with respect to a vector $r_{\alpha \gamma'}$, i.e., a vector from the pedestrian's position to the closest point on the vehicle's predicted paths.
A two-second constant velocity prediction is made along each lane the vehicle can follow. 

Similarly to the original model, we scale repulsive forces $F_{\alpha\beta}$ outside of a pedestrians field of view, i.e., 
\begin{align}
    F_{\alpha\beta} &= w(e_\alpha, -f_{\alpha\beta}) f_{\alpha\beta} \\
    F_{\alpha\gamma} &= w(e_\alpha, -f_{\alpha\gamma}) f_{\alpha\gamma}
\end{align}
with 
\begin{equation}
    w(e,f) = 
    \begin{cases}
        1 & \text{if $e \cdot f \geq \cos (\varphi), $} \\
        0.5 & \text{otherwise.}
    \end{cases}
\end{equation}

The final resulting force on a pedestrian $\alpha$ is thus
\begin{equation}
    F_\alpha = F_\alpha^0 + \sum_\beta F_{\alpha\beta} + \sum_\gamma F_{\alpha\gamma}
    \label{eq:completeForce}
\end{equation}
where we sum up all the repulsive social forces and vehicle forces, respectively. With the complete force model from \Cref{eq:completeForce}, we can update the state of each pedestrian with semi-implicit Euler integration such that
\begin{equation}
\begin{gathered}
    v_\alpha(t+\Delta t) = v_\alpha(t) + F_\alpha \Delta t \\ 
    p_\alpha(t+\Delta t) = p_\alpha(t) + v_\alpha(t+\Delta t) \Delta t \\  
\end{gathered}
\end{equation}
where $p(t)$, $v(t)$ are 2-dimensional position and velocity states at some time $t$.

\subsection{Pedestrian-aware Motion Planning}
\label{subsec:Pedestrian-awareMotionPlanning}

\begin{figure}[!htpb]
    \centering
    \input{Images/EvaluationFunnel.tex}
    \caption{Evaluation funnel for pedestrian-aware motion planning. This pipeline ensures safe and optimal trajectory selection by incorporating harm and risk assessments.}
    \label{fig:evaluation_funnel}
\end{figure}
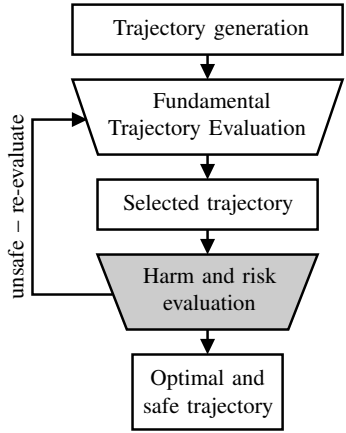

An existing motion planning algorithm~\cite{Frenetix} is used and enhanced accordingly, allowing for safety considerations specific to pedestrian interactions. As illustrated in \Cref{fig:evaluation_funnel}, a harm- and risk-based evaluation specifically for pedestrians is applied to further evaluate generated trajectories. The trajectories, which have already undergone feasibility checks and been assessed for comfort and efficiency using various cost functions, are now subjected to an additional safety evaluation. Trajectories that do not meet the defined safety thresholds are filtered out.

A key aspect of this process is the assessment of harm, which we define based on established frameworks in the literature. According to~\cite{Shiffrin2012}, harm encompasses various adverse effects on an individual's well-being, including physical and psychological impairments, as well as death. This definition aligns with widely accepted ethical principles, which prioritize preserving human life above all else, emphasizing the prevention of personal injury over property damage~\cite{BundesministeriumVerkehr2017}. 

To facilitate a quantifiable assessment of harm, the Abbreviated Injury Scale (AIS) is utilized, offering a standardized classification of injury severity~\cite{gennarelli2006ais}. The AIS scale ranges from 0, indicating no injury, to 6, representing a fatal injury. The AIS scores and their corresponding severity levels are summarized in \Cref{tab:ais_scores}.

\begin{table}[!htpb]
    \centering
    \caption{AIS scores of injury types.}
    \label{tab:ais_scores}
    \begin{tabularx}{0.8\linewidth}{c c c}
        \toprule
        AIS Score & Level of Severity & Description \\
        \midrule
        0 & No injury & Not injured \\
        1 & Minor & Superficial \\
        2 & Moderate & Reversible injuries \\
        3 & Serious & Reversible injuries \\
        4 & Severe & Life-threatening \\
        5 & Critical & Non-reversible injury \\
        6 & Fatal & Virtually not survivable \\
        \bottomrule
    \end{tabularx}
\end{table}

When individuals sustain multiple injuries, the Maximum AIS (MAIS) score is used to determine the overall severity, with the highest individual AIS score representing the MAIS.

To calculate the harm value, a logistic regression model is employed to estimate the probability of a severe injury occurring. Logistic regression is a statistical method that models the probability of a binary outcome~\cite{Kleinbaum2010}, which, in the present case, is the probability of an injury classified as MAIS3+ occurring. The general form of logistic regression is as follows:
\begin{equation}
P(Y=1) = \frac{1}{1 + e^{-(\beta_0 + \beta_1 X_1 + \beta_2 X_2 + \ldots + \beta_n X_n)}}
\end{equation}

\noindent In order to train the logistic regression model, data from the National Highway Traffic Safety Administration's Crash Report~\cite{NHTSA2024} is used, incorporating factors such as the mass $m$ of the involved objects, their relative speed $\Delta v$, and the collision angle $\alpha$. The harm value, denoted by $H(\xi)$, is then determined using the following equation:
\begin{equation}
H = \frac{1}{1 + e^{c_0 - c_1 \Delta v - c_{\text{area}}}}
\end{equation}
with
\begin{equation}
\Delta v = \frac{m_B}{m_A + m_B} \sqrt{v_A^2 + v_B^2 - 2 v_A v_B \cos \alpha}
\end{equation}
and empirically determined coefficients $c_0, c_1 \text{and } c_{\text{area}}$~\cite{geisslinger2021autonomous}.

Once the harm value $H(\xi)$ is established, the risk, denoted by $R(\xi)$, is defined as the product of the harm and the collision probability $p(\xi)$, as follows~\cite{geisslinger2021autonomous}:
\begin{equation}
R(\xi) = p(\xi) \cdot H(\xi)
\end{equation}

\begin{figure}[!ht]
    \centering
    \begin{tikzpicture}[font=\scriptsize]
        \node[inner sep=0pt] at (1.0,0) {\includegraphics[height=2.5mm]{Images/icons/ego.png}};
        \node[align=left, anchor=west] at (1.3,0) {Ego \\ vehicle};
    
        \node[inner sep=0pt] at (3.0,0) {\includegraphics[height=2.5mm]{Images/icons/dyn_obstacle.png}};
        \node[align=left, anchor=west] at (3.3,0) {Dynamic \\ obstacle};

        \node[inner sep=0pt] at (5.0,0) {\boxPedestrian};
        \node[align=left, anchor=west] at (5.1,0) {Pedestrians};

        \node[inner sep=0pt] at (7.0,0) {\UncertainPrediction};
        \node[align=left, anchor=west] at (7.2,0) {Uncertain \\ predictions};

        \node[anchor=west, inner sep=0pt] at (0.35,-1.4) {\includegraphics[width=0.95\linewidth, trim={14cm 14cm 14cm 12cm},clip]{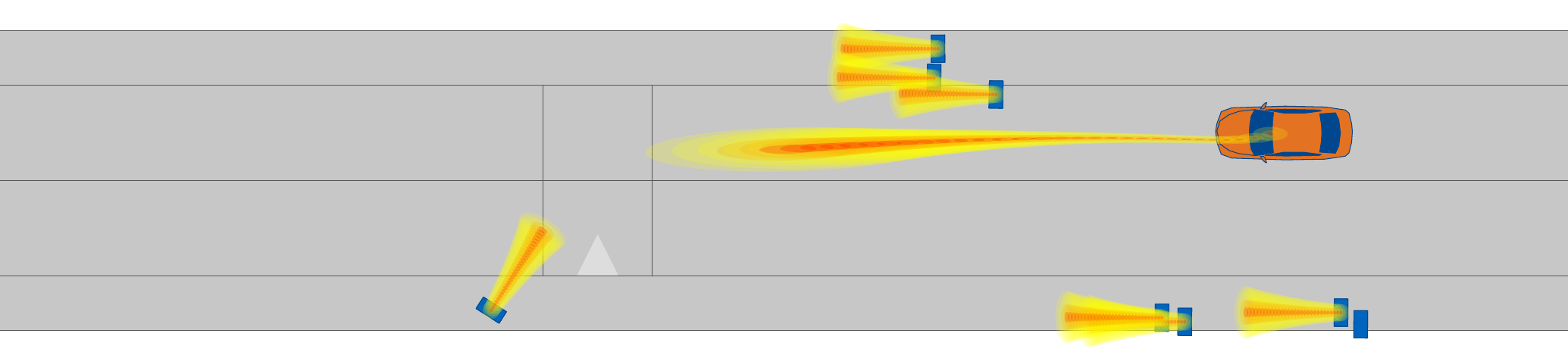}};

        \node[inner sep=0pt] at (1.1,-1.7) {\includegraphics[height=3.7mm]{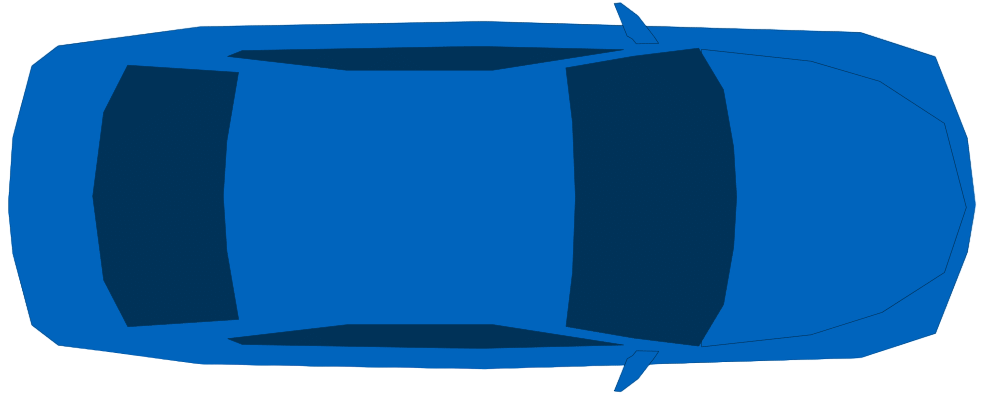}};
    
    \end{tikzpicture}
    \caption{Visualization of pedestrian and vehicle predictions with associated uncertainties. The uncertainty in the predicted positions grows over time, modeled by expanding ellipses.}
    \label{fig:UncertainPrediction}
\end{figure}

The collision probability $p(\xi)$ calculation relies on accurate pedestrian and vehicle predictions. In this work, predictions of other vehicles are generated using  Wale-Net~\cite{walenet}, a neural network-based prediction model that estimates the future positions of dynamic obstacles based on observed behaviors. The pedestrian predictions are computed using a constant velocity model that forecasts their future positions based on their current motion. These future positions are inherently uncertain and grow over time.
\Cref{fig:UncertainPrediction} visualizes these predictions, highlighting how the associated uncertainties expand as time progresses.

The uncertainties are mathematically represented using a bivariate normal distribution (BND). This probabilistic model captures the uncertainty in both the $ x $- and $ y $-coordinates of the predicted position. Formally, the predicted position $ \mathbf{X} $ of a pedestrian at any given time step can be expressed as~\cite{Henze2024}:
\begin{equation}
    \mathbf{X} \sim \mathcal{N}({\boldsymbol \mu}, {\boldsymbol \Sigma})
\end{equation}
where the mean vector $ {\boldsymbol \mu} $ and the symmetric, positive semi-definite covariance matrix $ {\boldsymbol \Sigma} $ are defined in \Cref{eq:meanandsigma}.
\begin{equation}
{\boldsymbol \mu} = 
\begin{pmatrix}
\mu_x \\
\mu_y
\end{pmatrix}, 
\quad
{\boldsymbol \Sigma} = 
\begin{pmatrix}
\sigma_x^2 & \sigma_{xy} \\
\sigma_{xy} & \sigma_y^2
\end{pmatrix}
\label{eq:meanandsigma}
\end{equation}
The mean vector $ {\boldsymbol \mu} $ represents the most likely position of the pedestrian in the $ x $- and $ y $-directions at a given time step, while the covariance matrix $ {\boldsymbol \Sigma} $ encodes the uncertainties associated with these predictions. Specifically, the diagonal elements $ \sigma_x^2 $ and $ \sigma_y^2 $ denote the variance in the pedestrian’s position along the $ x $- and $ y $-axes, respectively, indicating the degree of spread in each direction. The off-diagonal term $ \sigma_{xy} $ represents the covariance between the $ x $- and $ y $-coordinates, capturing the correlation between these uncertainties. The off-diagonal term can also be interpreted as the rotation of the uncertainty ellipse (UE)~\cite{Henze2024}.

To compute the collision probability $ p(\xi) $, we assess the likelihood that an obstacle's predicted position overlaps with the ego vehicle’s planned trajectory. We, therefore, calculate the probability that the obstacle’s probability mass, as described by the BND, lies within the region occupied by the ego vehicle. This probability can be expressed as an integral over the ego vehicle’s bounding box, defined by the interval $ [a, b] $ in the $ x $-direction and $ [c, d] $ in the $ y $-direction. The ego vehicle’s bounding box is intentionally enlarged to account for the spatial extent of pedestrians. The integral in \Cref{eq:pdfintegral} represents the total probability mass of the obstacle’s predicted position that falls within the vehicle’s space.
\begin{equation}
    p(\xi) = \int_{a}^{b} \int_{c}^{d} f_{X,Y}(x, y) \, dy \, dx
    \label{eq:pdfintegral}
\end{equation}
Here, $ f_{X, Y}(x, y) $ is the probability density function (PDF) of the BND. The PDF describes the relative likelihood of the obstacle being located at a specific position $ (x, y) $. While \Cref{eq:pdfintegral} provides an exact solution for the collision probability, solving it directly can be computationally expensive~\cite{Lambert2008}.

To avoid solving the integral, we employ the inclusion-exclusion principle, which approximates the collision probability by evaluating the cumulative distribution function (CDF) of the BND at the four corners $ (a, c), (a, d), (b, c), (b, d) $ of the ego vehicle’s bounding box. The CDF $ F_{X, Y}(x, y) $ represents the probability that the obstacle’s position lies within the region $ (-\infty, x] \times (-\infty, y] $, effectively giving the accumulated probability up to the point $(x, y)$. Using the inclusion-exclusion principle, the likelihood that the obstacle’s position falls within the bounding box can be computed as:
\begin{equation}
\begin{gathered}
    \tilde{p}(\xi) = P(a < X \leq b, c < Y \leq d) = \\ F_{X, Y}(b,d) - F_{X, Y}(a,d) - F_{X, Y}(b,c) + F_{X, Y}(a,c)
\end{gathered}
\end{equation}

\begin{figure}[!htpb]
    \centering
    \begin{tikzpicture}[font=\scriptsize]
        \node[anchor=west, inner sep=0pt] at (0.39,0) {\coloredbox{TUMBlue}};
        \node[align=left, anchor=west] at (0.75,0) {Bounding \\ box};

        \fill[black,opacity=1.0] (2.1,0) circle (0.07);
        \node[align=left, anchor=west] at (2.15,0) {Samples};

        \fill[GreenCR,opacity=1.0] (3.35,0) circle (0.07);
        \node[align=left, anchor=west] at (3.40,0) {Collision \\Samples};

        \fill[yellow,opacity=0.5] (4.8,0) ellipse (0.18 and 0.11);
        \node[align=left, anchor=west] at (4.9,0) {UE (3$\sigma$)};

        \fill[orange,opacity=0.7] (6.2,0) ellipse (0.14 and 0.07);
        \node[align=left, anchor=west] at (6.3,0) {UE (1$\sigma$)};

        \fill[red,opacity=0.7] (7.55,0) ellipse (0.08 and 0.04);
        \node[align=left, anchor=west] at (7.6,0) {UE (0.2$\sigma$)};

        \node[anchor=west, inner sep=0pt] at (0.35,-3.7) {\includegraphics[width=0.95\linewidth, trim={0cm 0cm 0cm 0cm},clip]{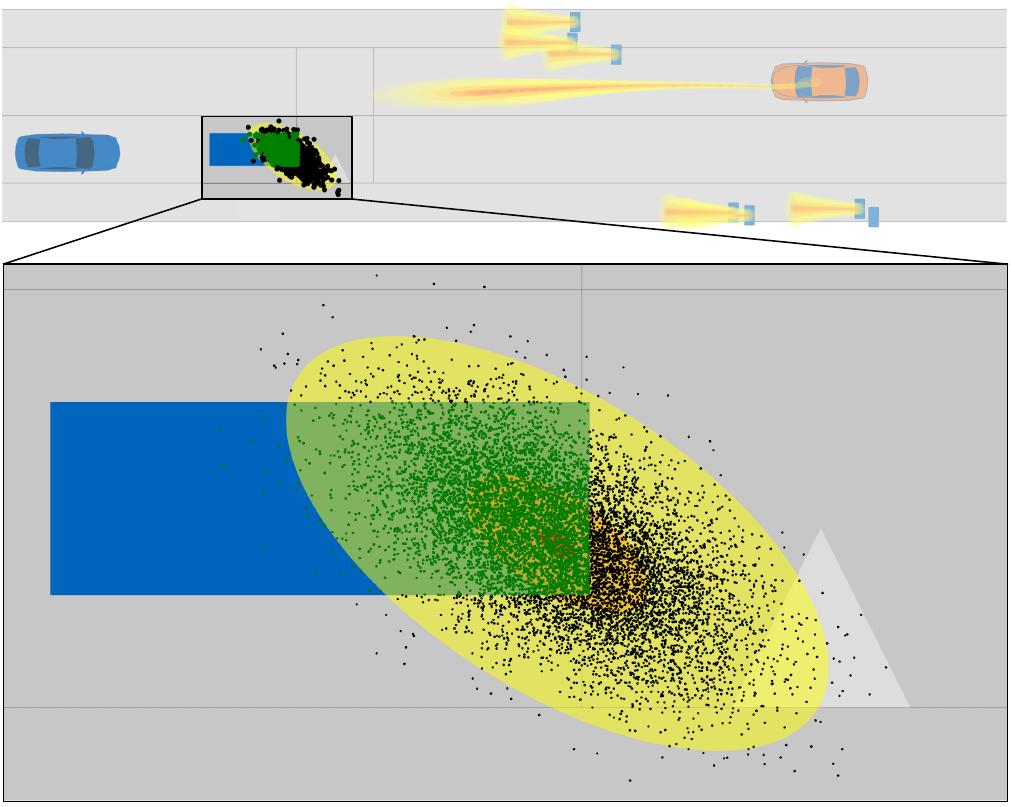}};
    
    \end{tikzpicture}
    \caption{Visualization of Monte Carlo sampling for collision probability estimation (here $p(\xi) = 0.5481$). The collision probability is estimated by determining the fraction of all sample points that fall within the ego vehicle's bounding box.}
    \label{fig:montecarlocollisionprob}
\end{figure}
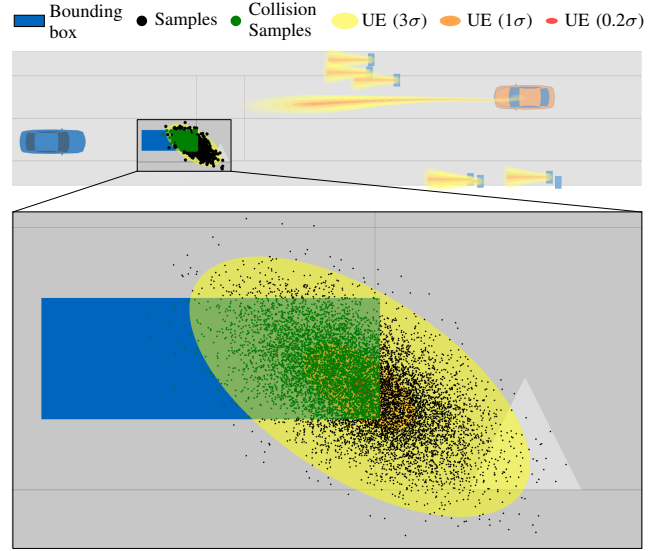

Our analytical approach can be numerically verified through Monte Carlo sampling. Monte Carlo sampling provides a complementary, intuitive approach for estimating this probability and offers clear visualizations, which help understand the distribution of predicted pedestrian positions relative to the ego vehicle's trajectory. In the Monte Carlo approach, random samples are drawn from the BND that represent the pedestrian’s predicted position. Each sample is checked whether it falls within the ego vehicle’s bounding box. The collision probability $\hat{p}(\xi)$ is then estimated by calculating the fraction of samples that lie within the vehicle’s space. \Cref{eq:montecarlosampling} estimates the probability $ \hat{p}(\xi) $.
\begin{equation}
\hat{p}(\xi) = \frac{1}{N} \sum_{i=1}^{N} \mathbf{1}_{\text{bbox}}(x_i, y_i)
\label{eq:montecarlosampling}
\end{equation}
where $ N $ is the number of samples, and 
\begin{equation}
\mathbf{1}_{\text{bbox}}(x_i, y_i) =
\begin{cases}
1 & \text{if } a \leq x_i \leq b \text{ and } c \leq y_i \leq d, \\
0 & \text{otherwise}.
\end{cases}
\end{equation}
This method, while computationally expensive for large-scale simulations, provides an effective way to numerically approximate the collision probability and can be visually represented, offering valuable insights into the spatial distribution of uncertainty.
A visualization of the Monte Carlo sampling approach is shown in \Cref{fig:montecarlocollisionprob}.

Although calculating collision probabilities and quantifying harm and risk values are technical tasks, addressing the associated risks raises important ethical considerations. Different approaches have been proposed in the literature, such as Bayes' rule for fair distribution of risk among traffic participants, the principle of equality, and the Maximin principle, which uses the maximum risk as a reference~\cite{geisslinger2021autonomous}. In this work, we adopt the Maximin principle. Using this approach, each trajectory is evaluated against the harm and risk threshold to determine if it meets safety requirements. The threshold values for harm $H_{\text{max}}$ and risk $R_{\text{max}}$ are not defined in this work, as they involve ethical decisions that are beyond the scope of this research.

Ultimately, a trajectory $\xi$ is considered valid or invalid based on whether it satisfies the safety requirements:
\begin{equation}
v_\xi = \ 
\begin{cases}
\text{valid}, & \text{if } H(\xi) < H_{\text{max}}, R(\xi) < R_{\text{max}}  \\
\text{invalid}, & \text{otherwise}
\end{cases}
\end{equation}


\section{Results}


\subsection{Simulation Setup}

Both the motion planner and the pedestrian simulator are deployed in a Commonroad simulation environment~\cite{althoff_commonroad_2017} using modified scenarios (as described in \Cref{subsec:PedestrianSimulator}). 
For comparison, a baseline planner~\cite{Frenetix} is executed with consistent settings, without scenario-specific adjustments or fine-tuning. This approach ensures that the results remain unbiased and independent of scenario-specific optimizations. Our approach builds upon this baseline by incorporating the risk-aware module. To analyze computation times, simulations were conducted on a machine with an AMD Ryzen 9 processor featuring 16 cores running at a base clock frequency of 4.5 GHz with 128 GB of RAM. The operating system used was Ubuntu 22.04.

\subsection{Pedestrian Simulation}
\label{subsec:result-pedestriansimulatoin}

\begin{figure}[!htpb]
    \centering
    \includegraphics[width =0.85\linewidth, trim={11.5cm 3.5cm 10.5cm 4cm}, clip]{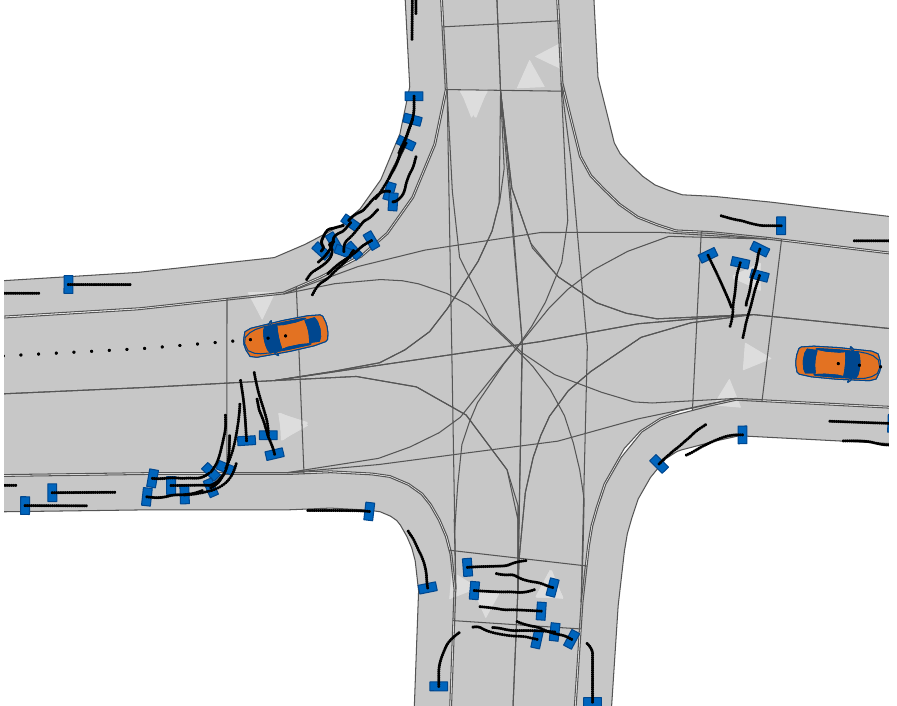}
    \caption{Simulation of pedestrian trajectories at a busy intersection. More than 300 pedestrians are simulated, with future movements shown over the next three seconds.}
    \label{fig:ped_sim_ex}
\end{figure}

We evaluate the computational performance of our pedestrian simulator by analyzing two key aspects: offline policy calculation time and simulation step duration. To assess the efficiency of offline policy computation, we measured the time required to create sets of policies under varying numbers of goal points across scenarios with different levels of complexity. The scenarios include a straight road (Scenario 1), a T-junction (Scenario 2), and a 4-way intersection (Scenario 3, see \Cref{fig:ped_sim_ex}), each progressively larger and more complex. \Cref{tab:offline_policy_time_scenarios} presents the offline policy calculation time, illustrating how runtime scales with the number of goal points and scenario complexity.
\begin{table}[ht]
    \centering
    \caption{Offline policy calculation time for different scenarios}
    \begin{tabularx}{0.75\linewidth}{c c c}
    \toprule
    \textbf{Scenario} & \textbf{Goal Points} & \textbf{Calculation Time in \si{\second}} \\
    \midrule
    \multirow{3}{*}{1: \SI{2299}{\square\meter}} 
      & 1 & 0.79 \\
      & 2 & 0.80 \\
      & 8 & 0.87 \\
    \midrule
    \multirow{3}{*}{2: \SI{5153}{\square\meter}} 
      & 1 & 6.81 \\
      & 2 & 7.05 \\
      & 8 & 33.29 \\
    \midrule
    \multirow{3}{*}{3: \SI{6867}{\square\meter}} 
      & 1 & 12.56 \\
      & 2 & 13.84 \\
      & 8 & 88.53 \\
    \bottomrule
    \end{tabularx}
    \label{tab:offline_policy_time_scenarios}
\end{table}

\noindent The results show that while computation time generally increases with both scenario complexity and the number of goal points, the increase is not strictly linear due to parallel processing optimizations. For instance, for 8 goal points, Scenario 1 required only \SI{0.87}{\second}, whereas Scenario 2 took \SI{33.29}{\second}, highlighting the impact of scenario complexity on computational demand.

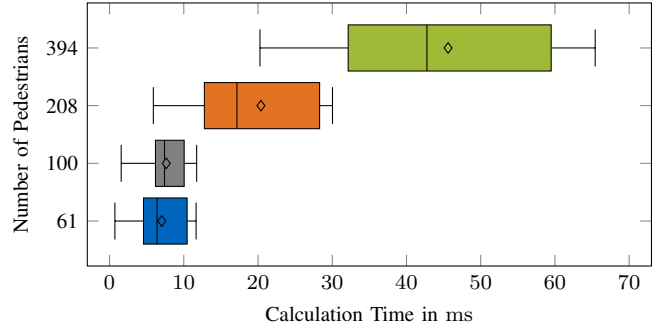
\begin{figure}[!ht]
    \centering
    \input{Images/simulation_step_duration.tex}
    \caption{Simulation step durations for advancing scenario 3 by one step. The boxplot illustrates the calculation times for different numbers of pedestrians. The mean \( \mu \) is marked as a diamond and the median \( \tilde{x} \) is depicted as a horizontal line within each box.}
    \label{fig:simulation_step_time}
\end{figure}

In addition to offline policy calculation, we assessed the simulation step duration, representing the time required to advance all pedestrians by one step in the simulation. Although real-time feasibility is not the primary focus, step duration is critical for ensuring efficient large-scale simulations. For Scenario 3, we varied the number of pedestrians and measured the step times over approximately 100 simulation steps. The statistical evaluation of the pedestrian simulator step duration, including the mean \( \mu \) and median \( \tilde{x} \), is depicted in \Cref{fig:simulation_step_time}.

The results demonstrate that step durations increase with the number of pedestrians but remain stable and suitable for real-time simulation within the evaluated range.

To complement the runtime analysis, we conducted a qualitative evaluation using Scenario 3. \Cref{fig:ped_sim_ex} illustrates the simulated trajectories of pedestrians over a three-second interval, revealing how they progress toward their goals while dynamically interacting with each other. The visualization highlights pedestrians’ preference for SWs and designated crossings, as well as their ability to avoid collisions through adaptive behaviors.

\begin{figure}[!ht]
    \centering
    \includegraphics[width=0.8\linewidth, trim={8cm 4cm 7cm 5cm}, clip]{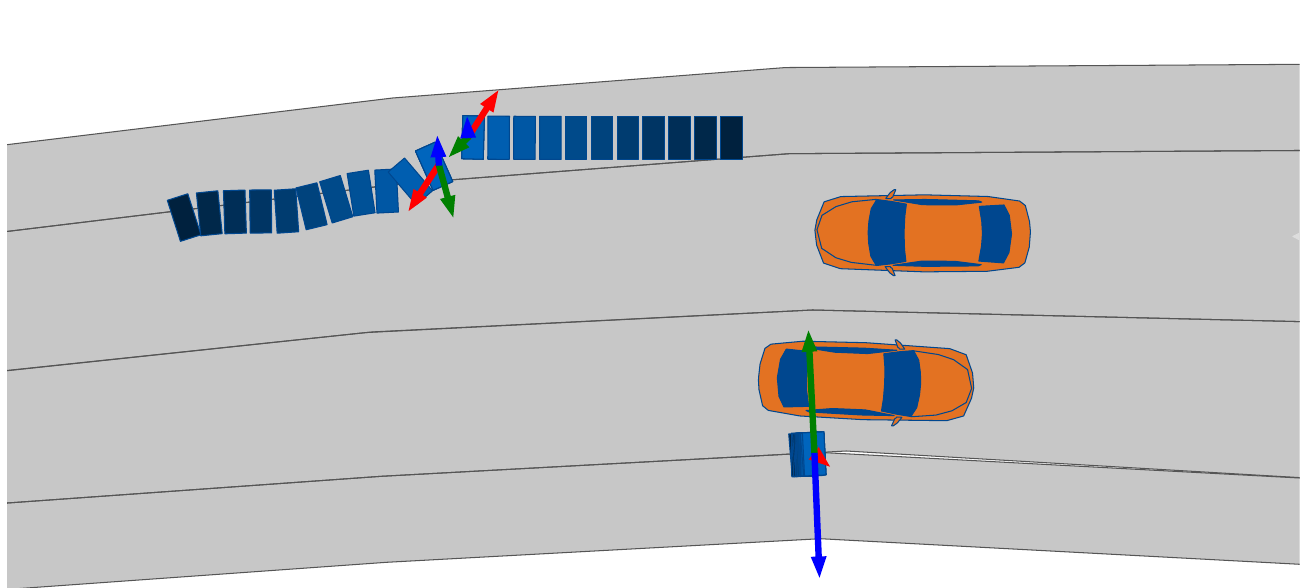}
    \caption{Previous pedestrian positions visualized in progressively darker shades of blue, representing \SI{0.5}{\second} intervals up to \SI{5}{\second} in the past.}
    \label{fig:ped_sim_forces}
\end{figure}

\Cref{fig:ped_sim_forces} further illustrates pedestrian behavior in response to surrounding influences, including repelling forces from vehicles and social interactions between pedestrians. The visualization of previous pedestrian positions, shown in progressively darker shades of blue, provides insight into their adaptive behaviors as they balance goal-oriented movement with collision avoidance.
In the lower part of the figure, the behavior of one pedestrian demonstrates a waiting action as it stops to let a vehicle pass, responding to the repelling force exerted by the vehicle. In the upper part, another pedestrian veers away from the street as it approaches a nearby vehicle. Here, the vehicle’s repelling force exceeds the social force exerted by another pedestrian, prompting a change in trajectory.


\subsection{Motion Planning in pedestrian-rich Scenarios}
\label{subsec:result-planning}

In this section, we analyze the computational performance, vehicle behavior, and overall effectiveness of our proposed pedestrian-aware motion planner. The analysis includes runtime performance, qualitative assessments in specific scenarios, and a quantitative comparison of different planner configurations using various scenarios.

The runtime analysis, including the mean \( \mu \), median \( \tilde{x} \), and standard deviation \( \sigma \), is shown in \Cref{tab:harm_risk_time}. It reflects the computational demand required to assess safety per trajectory. As the number of visible pedestrians near the ego vehicle increases, the computation time grows significantly.
\begin{table}[ht]
\centering
\caption{Safety Assessment Runtime}
\begin{tabularx}{0.8\linewidth}{c c c c}
    \toprule
    Number of Pedestrians & $\mu$ in \si{\milli\second} & $\tilde{x}$ in \si{\milli\second} & $\sigma$ in \si{\milli\second} \\
    \midrule
    1  & 6.98 & 7.04 & 0.15 \\ 
    10 & 64.64 & 64.64 & 0.48 \\ 
    20 & 133.99 & 134.79 & 1.68 \\ 
    \bottomrule
    \end{tabularx}
\label{tab:harm_risk_time}
\end{table}
As illustrated, the average runtime remains relatively low with a single pedestrian, requiring only about \SI{6.98}{\milli\second} per trajectory check. However, as the number of pedestrians increases to 20, the computation time rises sharply to an average of \SI{133.99}{\milli\second}. These values are averaged over 100 trials per configuration, ensuring statistically robust results. This increase in computation time is primarily because safety assessments need to be performed for each pedestrian in the vicinity of the ego vehicle. While the per-trajectory assessment times provide a baseline, the total computation time required to find a feasible trajectory can be significantly higher. The planner must repeatedly evaluate candidate trajectories until one meets the risk thresholds. As a result, the overall computation time increases, especially in scenarios where many trajectories must be filtered out before identifying a safe and viable option.

\begin{figure*}[!htpb]
    \centering
    \begin{tikzpicture}[font=\footnotesize,trim axis right]



        \node[inner sep=0.5pt, anchor=south west, draw=black, dashed, line width=1pt] (img1) at (0.0, 3.7) {\includegraphics[width=0.42\linewidth, trim={20cm 4cm 20cm 25cm},clip]{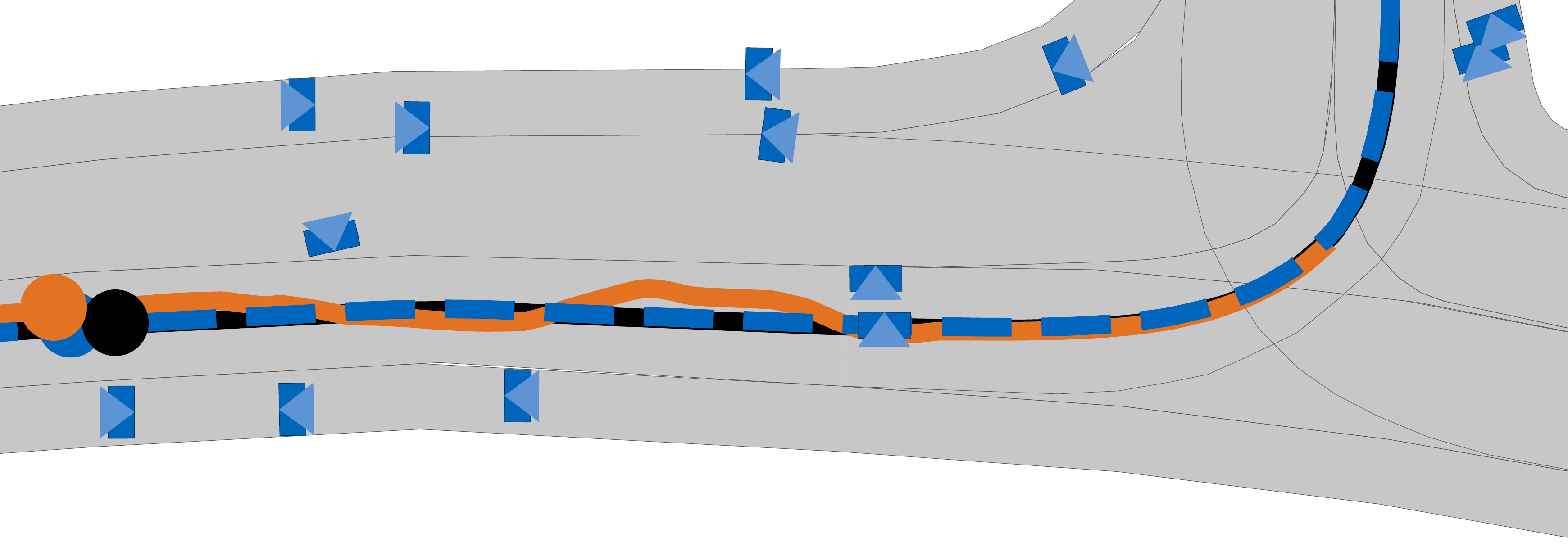}};

        \node[inner sep=0.5pt, anchor=south east, draw=black, dashed, line width=1pt] (img2) at (16.0, 3.7) {\includegraphics[width=0.42\linewidth, trim={20cm 4cm 20cm 25cm},clip]{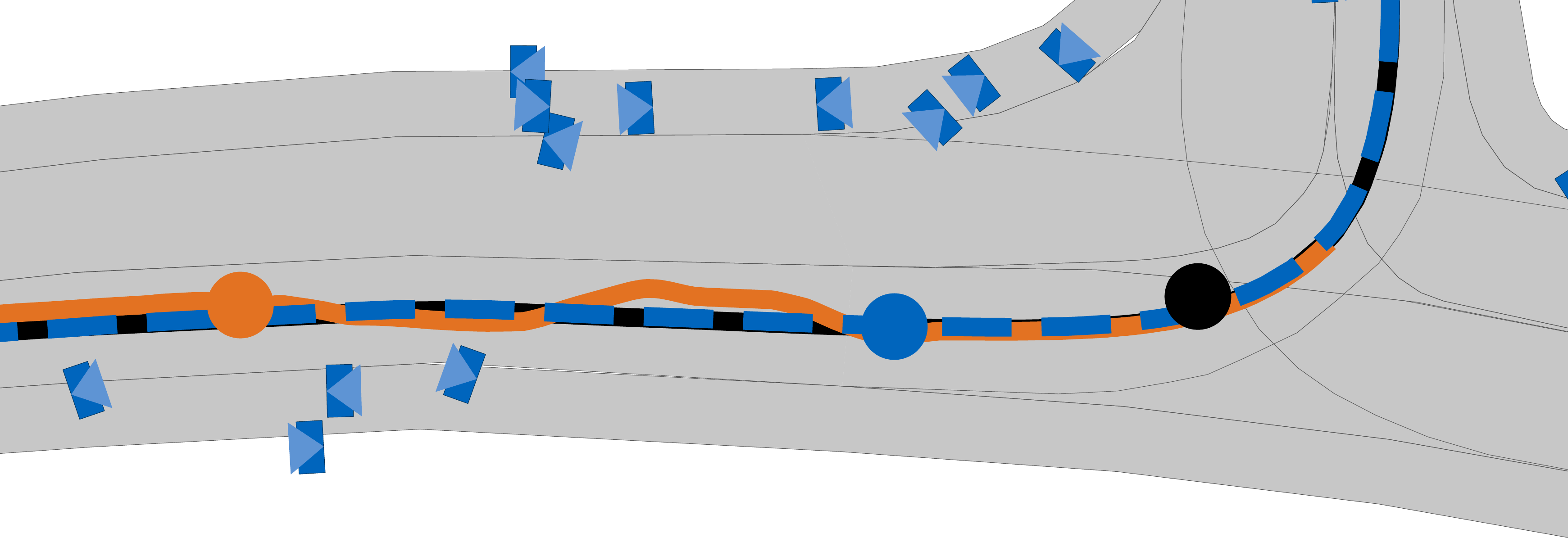}};

        \input{Images/scenario2_risk}

        \draw[dashed, line width=1pt, black] (img1.south) -- (3.45, 3.3);

        \draw[dashed, line width=1pt, black] (img2.south) -- (10.3, 3.3);
    \end{tikzpicture}
    
    \vspace{0.1cm}
    
    \input{Images/scenario2_velocity}

    \caption{Risk $R$ and velocity $v$ profiles for different planner configurations in Scenario 2, illustrating the impact of varying risk thresholds on vehicle behavior. The upper bird's-eye view (BEV) visualizations, using $R_\mathrm{max} = 0.075$ for the risk-aware planner, show two timesteps with ego-vehicle and selected (enlarged) pedestrian positions for clarity.}

    \label{fig:scenario2_qualitative_analysis}
\end{figure*}

We further examined the vehicle’s qualitative behavior in Scenario 2, which was populated with 315 pedestrians under different planner settings. We tested three configurations: (1) our proposed risk-aware planner with adjustable thresholds, (2) the baseline collision-probability-only planner, and (3) an aggressive, non-cautious planner that prioritizes efficiency without regard for pedestrian safety. The results, visualized in \Cref{fig:scenario2_qualitative_analysis}, reveal distinct behaviors across configurations. Notably, the risk peaks are shifted in time as the different planner configurations encounter critical areas of the scenario at different moments.

For our risk-aware planner, the vehicle adjusts its speed to maintain the predefined acceptable risk level without becoming overly cautious. As risk thresholds decrease, we observe lower maximum speeds, especially near pedestrians, while allowing the vehicle to progress consistently. Notably, the planner does not resort to immediate full braking when pedestrians unexpectedly enter the roadway, especially during illegal crossings. Instead, it adjusts speed gradually, ensuring pedestrian safety while maintaining vehicle flow. 

In contrast, the aggressive planner prioritizes speed and rapid progress, tolerating higher risk levels and potentially exposing VRUs to greater danger. The baseline approach often stalls in anticipation of illegal pedestrian crossings, allowing pedestrians to cross without intervention. While reducing risk, this cautious behavior limits the vehicle’s operational capacity in scenarios with high pedestrian density, resulting in extended stops and an increased likelihood of freezing. However, in certain cases, the baseline planner exhibited counterintuitive behavior by accelerating to pass before a pedestrian to avoid collisions. While this approach may preserve motion, it compromises safety near VRUs, highlighting the need for better control over trajectory selection in such situations.

\begin{figure}[!ht]
    \centering
    \input{Images/scenario3_crosswalk_velocity_trajectory_number}
    \caption{Speed and time profiles in Scenario 1 with a crosswalk (shaded area). The BEV shows vehicle positions at $t = 95$. The bottom plot depicts velocity and time progression along the traveled distance.}
    \label{fig:scenario1_speed_profiles}
\end{figure}
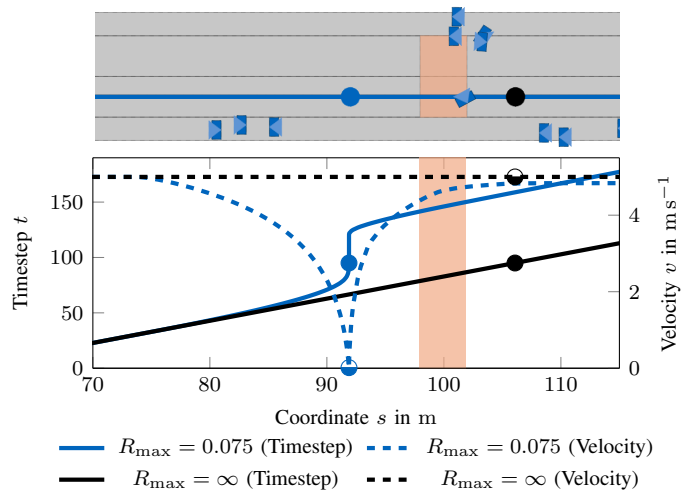

To further assess the planner’s adaptability, we analyzed a variation of Scenario 1 with an additional pedestrian CW. In this setup, we compared the aggressive planner and the risk-aware planner, which integrates crosswalk detection capabilities. \Cref{fig:scenario1_speed_profiles} visualizes the results.

Our risk-aware planner appropriately slows down near the crosswalk and, if necessary, stops completely to allow pedestrians to cross. The progress plot shows that the planner waits several seconds until all pedestrians have fully cleared the crosswalk, demonstrating its cautious behavior. In contrast, the aggressive planner disregards the presence of the crosswalk entirely, maintaining speed and driving through without stopping. This is reflected in its speed profile, where no deceleration is observed. Consequently, pedestrians are denied their right of way and are exposed to higher risks.

We conducted a quantitative evaluation to benchmark the three planner configurations to complete the analysis, with a moderate threshold set at $R_\mathrm{max}=0.075$ for our risk-aware planner. We executed 100 simulation runs for each configuration using a standardized scenario with varying pedestrian configurations, densities, and initial positions. Each simulation ran for 100 timesteps, measuring efficiency through average, minimum, and maximum distance traveled and speed values. In addition, risk metrics were measured, as shown in \Cref{tab:quantitative_comparison}.
\begin{table}[ht]
\centering
\caption{Quantitative Comparison of Planner Configurations}
\begin{tabularx}{\linewidth}{c c c c}
    \toprule
    Planner & \parbox[c]{2.1cm}{\centering Traveled Distance \\ in \si{\meter}} & Risk & Velocity in \si{\meter\per\second} \\
    \midrule
    \multirow{3}{*}{Baseline} & Mean: $41.54 $& Mean: $0.002$ & Mean: $4.008$ \\
                               & Min $11.16$ & Min: $0.000$  & Min: $0.006$ \\
                               & Max $57.36$ & Max: $0.092$  & Max: $8.729$ \\
    \midrule
    \multirow{3}{*}{\parbox[c]{1.8cm}{\centering Aggressive\\$R_\mathrm{max}=\infty$}} & Mean $53.60$ & Mean: $0.045$ & Mean: $5.517$ \\
                                 &  Min: $39.70$ & Min: $0.000$  & Min: $4.717$ \\
                                 &  Max: $54.21$ & Max: $0.165$  & Max: $5.702$ \\
    \midrule
    \midrule
    \multirow{3}{*}{\parbox[c]{1.8cm}{\centering Risk-Aware\\$R_\mathrm{max}=7.5 \%$\\\textbf{ours}}} & Mean: $52.22$ & Mean: $0.029$ & Mean: $5.376$ \\
                                 & Min: $46.24$ & Min: $0.000$  & Min: $3.684$ \\
                                 & Max: $53.77$ & Max: $0.075$  & Max: $5.630$ \\
    \bottomrule
\end{tabularx}
\label{tab:quantitative_comparison}
\end{table}
The quantitative evaluation reveals that the baseline planner consistently results in shorter maximum travel distances accompanied by lower average speeds. In several configurations, the vehicle covers only minimal distances, with its speed reduced to nearly zero. This indicates that the vehicle frequently stops to allow pedestrians to pass, ultimately failing to continue its journey within the simulation time. Additionally, the planner occasionally opts for higher speeds to avoid collisions, as already seen in the qualitative evaluation. In contrast, both the aggressive planner and our risk-aware approach achieve higher travel distances while maintaining moderate speeds. The aggressive planner, however, encounters a collision, resulting in a shorter travel distance in one configuration. It also tolerates higher risk levels. Our risk-aware planner successfully balances efficiency and safety. It maintains moderate speeds while ensuring risk levels do not exceed the predefined threshold.


\section{Discussion}
Developing an adaptive, pedestrian-aware motion planner addresses a fundamental challenge in AV navigation within complex urban environments. Our approach integrates risk considerations into the decision-making process, enabling the planner to balance safety and efficiency. Results show that our pedestrian-aware planner provides substantial improvements over traditional planners that focus solely on collision probabilities.

A central component of this framework is the pedestrian simulation model, which serves as a robust foundation for developing and testing motion planning algorithms in a dynamic and interactive environment. By modeling pedestrian behavior, the simulator allows AVs to interact with pedestrians, making it possible to evaluate planner performance in urban scenarios with high-density and unpredictable flows. However, a current limitation is that the simulator applies uniform behavior modeling across all pedestrians without accounting for differences in pedestrian behavior. This limits the ability to simulate behaviors that vary by culture or locale.

The simulator’s high number of objects highlights the computational demands on the motion planner itself, emphasizing the importance of runtime efficiency. As the planner evaluates and filters multiple trajectory options, the density of surrounding pedestrians and other objects can quickly escalate computational load. Meeting the necessary latency for real-world applications in these scenarios proves challenging, as the planner must process frequent and close interactions with pedestrians without compromising performance.

Nevertheless, our findings demonstrate that safe navigation in pedestrian-rich scenarios can be achieved by carefully setting risk thresholds. However, selecting appropriate thresholds remains complex and context-dependent, potentially requiring regulatory input and adaptive adjustments. It is important to note that all experiments conducted thus far were performed in a simulation environment. Future work will involve validating these approaches in real-world scenarios, including tests on an actual vehicle,


\section{Conclusion \& Outlook}

In this paper, we presented a pedestrian-aware motion planning algorithm explicitly designed to address the unique challenges posed by dense urban environments. Our approach combines a risk-aware motion planner with a pedestrian simulator capable of producing realistic, interactive pedestrian behavior. By integrating social force principles within the pedestrian simulator, we created a dynamic testing environment that enables AVs to respond to crowded and complex scenarios. By simulating different scenarios, we evaluated various configurations of risk thresholds to balance safety and efficiency and assessed the impact of these settings on the AV’s behavior.

The results of our experiments reveal several key insights: First, the careful tuning of risk thresholds allows the AV to navigate pedestrian-rich areas without overly conservative or aggressive behavior, enabling safe and efficient motion. Additionally, the results highlight the computational demands associated with real-time trajectory evaluation, particularly in scenarios with a high density of pedestrians and other interacting objects.

For future research, several promising directions exist to further enhance both the pedestrian simulation model and the pedestrian-aware motion planner. For example, improving the simulator’s modeling capabilities by incorporating region-specific pedestrian behavior or group dynamics could enable more nuanced interactions and facilitate development for diverse urban settings. Additionally, expanding the planner's scalability and performance is critical. To this end, integrating the safety assessment directly into the C++ codebase, rather than relying on external processes, could improve runtime efficiency. Parallel computation of multiple trajectories would also help minimize delays and avoid computational bottlenecks. 

Further enhancements could include dynamically adjusting risk thresholds in response to environmental factors or evaluating a combination of safety metrics to guide trajectory selection, allowing for more flexible and contextually responsive decision-making. These improvements would contribute to a more robust, adaptable, and reliable motion planning system.


\section*{Acknowledgement}
As the first author, Korbinian Moller initiated the idea of this article and significantly contributed to its conceptualization, implementation, and content. Truls Nyberg provided valuable contributions, particularly in developing and implementing the pedestrian simulation model, and actively participated in the analysis and manuscript revision. Jana Tumova and Johannes Betz critically reviewed the article, approved the final version to be published, and supported the research project. The authors would like to thank the Munich Institute of Robotics and Machine Intelligence (MIRMI) for their continuous support.


\bibliographystyle{IEEEtran}
\bibliography{bibliography}

\begin{IEEEbiography}[{\includegraphics[width=1in,height=1.25in, clip,keepaspectratio]{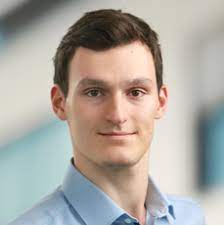}}]{Korbinian Moller} received a B.Sc. degree and an M.Sc. degree in mechanical engineering from the Technical University of Munich (TUM) in 2021 and 2023, respectively. He is currently pursuing a Ph.D. degree at the Professorship of Autonomous Vehicle Systems (AVS) at TUM. His research interests include edge-case scenario simulation, the optimization of vehicle behavior, and motion planning in autonomous driving.
\end{IEEEbiography}

\vskip -1\baselineskip 

\begin{IEEEbiography}[{\includegraphics[width=1in,height=1.25in, clip,keepaspectratio]{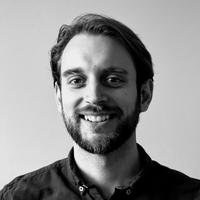}}]{Truls Nyberg} is currently pursuing a Ph.D. with a joint affiliation at Scania CV AB and the Division of Robotics, Perception, and Learning at the KTH Royal Institute of Technology in Stockholm, Sweden, with partial support from the Wallenberg AI, Autonomous Systems, and Software Program (WASP), funded by the Knut and Alice Wallenberg Foundation. As part of his studies, he completed a research visit at the Autonomous Vehicle Systems (AVS) Lab at the Technical University of Munich. His research focuses on risk-aware decision-making and situational awareness for autonomous vehicles, emphasizing applications for heavy-duty trucks and buses. He received his M.Sc. in Engineering Applied Physics and Electrical Engineering from Linköping University in 2018, specializing in Control and Information Systems.
\end{IEEEbiography}

\vskip -1\baselineskip 

\begin{IEEEbiography}[{\includegraphics[width=1in,height=1.25in, clip,keepaspectratio]{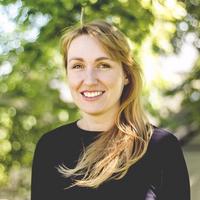}}]{Jana Tumova} is an Associate Professor at the Division of Robotics, Perception and Learning at KTH Royal Institute of Technology. She holds a Ph.D. in computer science from Masaryk University and received an ACCESS postdoctoral fellowship at KTH in 2013. She has also been a visiting researcher at MIT, Boston University, and the Singapore-MIT Alliance. Her research interests include formal methods for decision-making, motion planning, and control in autonomous systems. She received the Swedish Research Council Starting Grant and the Early Career Award from the Robotics: Science and Systems Foundation.
\end{IEEEbiography}

\vskip -1\baselineskip 

\begin{IEEEbiography}[{\includegraphics[width=1in,height=1.25in, clip,keepaspectratio]{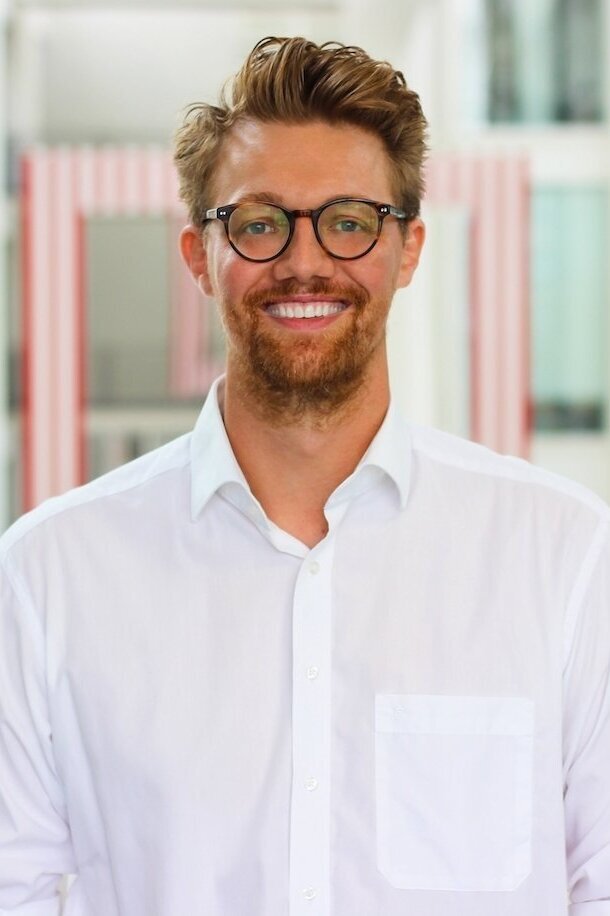}}]{Johannes Betz} is an assistant professor in the Department of Mobility Systems Engineering at the Technical University of Munich (TUM). He is one of the founders of the TUM Autonomous Motorsport team. His research focuses on developing adaptive dynamic path planning and control algorithms, decision-making algorithms that work under high uncertainty in multi-agent environments, and validating the algorithms on real-world robotic systems. Johannes earned a B.Eng. (2011) from the University of Applied Science Coburg, an M.Sc. (2012) from the University of Bayreuth, an M.A. (2021) in philosophy from TUM, and a Ph.D. (2019) from TUM. 
\end{IEEEbiography}

\vfill

\end{document}

%% file: Images/Framework.tex
\resizebox{0.85\textwidth}{!}{
\tikzset{every picture/.style={line width=1.0pt}} 

\begin{tikzpicture}[x=1.0pt,y=0.8pt,yscale=-1,xscale=1]

\draw  [color={rgb, 255:red, 0; green, 0; blue, 0 }  ,draw opacity=1 ][fill={rgb, 255:red, 230; green, 230; blue, 230 }  ,fill opacity=1 ] (145,135) -- (525,135) -- (525,360) -- (145,360) -- cycle ;
\draw  [dashed, color={rgb, 255:red, 0; green, 0; blue, 0 }  ,draw opacity=1 ](110,80) -- (540,80) -- (540,375) -- (110,375) -- cycle ;
\draw    (100,105) -- (142,105) ;
\draw [shift={(145,105)}, rotate = 180] [fill={rgb, 255:red, 0; green, 0; blue, 0 }  ][line width=0.08]  [draw opacity=0] (6.25,-3) -- (0,0) -- (6.25,3) -- cycle    ;
\draw   (145,90) -- (255,90) -- (255,120) -- (145,120) -- cycle ;
\draw  [color={rgb, 255:red, 0; green, 0; blue, 0 }  ,draw opacity=1 ][fill={rgb, 255:red, 255; green, 255; blue, 255 }  ,fill opacity=1 ] (160,180) -- (310,180) -- (310,340) -- (160,340) -- cycle ;
\draw   (280,90) -- (390,90) -- (390,120) -- (280,120) -- cycle ;
\draw    (255,105) -- (277,105) ;
\draw [shift={(280,105)}, rotate = 180] [fill={rgb, 255:red, 0; green, 0; blue, 0 }  ][line width=0.08]  [draw opacity=0] (6.25,-3) -- (0,0) -- (6.25,3) -- cycle    ;
\draw   (415,90) -- (525,90) -- (525,120) -- (415,120) -- cycle ;
\draw    (390,105) -- (412,105) ;
\draw [shift={(415,105)}, rotate = 180] [fill={rgb, 255:red, 0; green, 0; blue, 0 }  ][line width=0.08]  [draw opacity=0] (6.25,-3) -- (0,0) -- (6.25,3) -- cycle    ;
\draw  [color={rgb, 255:red, 0; green, 0; blue, 0 }  ,draw opacity=1 ][fill={rgb, 255:red, 255; green, 255; blue, 255 }  ,fill opacity=1 ] (360,180) -- (510,180) -- (510,340) -- (360,340) -- cycle ;
\draw  [fill={rgb, 255:red, 255; green, 255; blue, 255 }  ,fill opacity=1 ] (275,140) -- (395,140) -- (395,170) -- (275,170) -- cycle ;
\draw    (470,120) -- (470,150) -- (398,150) ;
\draw [shift={(395,150)}, rotate = 360] [fill={rgb, 255:red, 0; green, 0; blue, 0 }  ][line width=0.08]  [draw opacity=0] (6.25,-3) -- (0,0) -- (6.25,3) -- cycle    ;
\draw   (175,220) -- (295,220) -- (295,250) -- (175,250) -- cycle ;
\draw    (275,160) -- (235,160) -- (235,177) ;
\draw [shift={(235,180)}, rotate = 270] [fill={rgb, 255:red, 0; green, 0; blue, 0 }  ][line width=0.08]  [draw opacity=0] (6.25,-3) -- (0,0) -- (6.25,3) -- cycle    ;
\draw    (395,160) -- (435,160) -- (435,177) ;
\draw [shift={(435,180)}, rotate = 270] [fill={rgb, 255:red, 0; green, 0; blue, 0 }  ][line width=0.08]  [draw opacity=0] (6.25,-3) -- (0,0) -- (6.25,3) -- cycle    ;
\draw  [color={rgb, 255:red, 0; green, 0; blue, 0 }  ,draw opacity=1 ][fill={rgb, 255:red, 155; green, 155; blue, 155 }  ,fill opacity=1 ] (160,180) -- (310,180) -- (310,210) -- (160,210) -- cycle ;
\draw  [color={rgb, 255:red, 0; green, 0; blue, 0 }  ,draw opacity=1 ][fill={rgb, 255:red, 155; green, 155; blue, 155 }  ,fill opacity=1 ] (360,180) -- (510,180) -- (510,210) -- (360,210) -- cycle ;
\draw   (175,260) -- (295,260) -- (295,290) -- (175,290) -- cycle ;
\draw   (175,300) -- (295,300) -- (295,330) -- (175,330) -- cycle ;
\draw    (235,210) -- (235,217) ;
\draw [shift={(235,220)}, rotate = 270] [fill={rgb, 255:red, 0; green, 0; blue, 0 }  ][line width=0.08]  [draw opacity=0] (6.25,-3) -- (0,0) -- (6.25,3) -- cycle    ;
\draw    (235,250) -- (235,257) ;
\draw [shift={(235,260)}, rotate = 270] [fill={rgb, 255:red, 0; green, 0; blue, 0 }  ][line width=0.08]  [draw opacity=0] (6.25,-3) -- (0,0) -- (6.25,3) -- cycle    ;
\draw    (235,290) -- (235,297) ;
\draw [shift={(235,300)}, rotate = 270] [fill={rgb, 255:red, 0; green, 0; blue, 0 }  ][line width=0.08]  [draw opacity=0] (6.25,-3) -- (0,0) -- (6.25,3) -- cycle    ;
\draw   (375,220) -- (495,220) -- (495,250) -- (375,250) -- cycle ;
\draw   (375,260) -- (495,260) -- (495,290) -- (375,290) -- cycle ;
\draw   (375,300) -- (495,300) -- (495,330) -- (375,330) -- cycle ;
\draw    (435,210) -- (435,217) ;
\draw [shift={(435,220)}, rotate = 270] [fill={rgb, 255:red, 0; green, 0; blue, 0 }  ][line width=0.08]  [draw opacity=0] (6.25,-3) -- (0,0) -- (6.25,3) -- cycle    ;
\draw    (435,250) -- (435,257) ;
\draw [shift={(435,260)}, rotate = 270] [fill={rgb, 255:red, 0; green, 0; blue, 0 }  ][line width=0.08]  [draw opacity=0] (6.25,-3) -- (0,0) -- (6.25,3) -- cycle    ;
\draw    (435,290) -- (435,297) ;
\draw [shift={(435,300)}, rotate = 270] [fill={rgb, 255:red, 0; green, 0; blue, 0 }  ][line width=0.08]  [draw opacity=0] (6.25,-3) -- (0,0) -- (6.25,3) -- cycle    ;
\draw    (235,330) -- (235,350) -- (335,350) ;
\draw    (435,330) -- (435,350) -- (335,350) ;
\draw [shift={(335,350)}, rotate = 180] [color={rgb, 255:red, 0; green, 0; blue, 0 }  ][fill={rgb, 255:red, 0; green, 0; blue, 0 }  ][line width=0.75]      (0, 0) circle [x radius= 2.34, y radius= 2.9]   ;
\draw    (335,350) -- (335,368) -- (130,368) -- (130,150) -- (272,150) ;
\draw [shift={(275,150)}, rotate = 180] [fill={rgb, 255:red, 0; green, 0; blue, 0 }  ][line width=0.08]  [draw opacity=0] (6.25,-3) -- (0,0) -- (6.25,3) -- cycle    ;

\draw (157,99) node [anchor=north west][inner sep=0.75pt]   [align=left] {{\small Scenario preprocessing}};
\draw (111,67) node [anchor=north west][inner sep=0.75pt]   [align=left] {Simulation Environment};
\draw (297,93) node [anchor=north west][inner sep=0.75pt]   [align=left] {{\small Pedestrian simulator }};
\draw (313,106) node [anchor=north west][inner sep=0.75pt]   [align=left] {{\small initialization}};
\draw (439,93) node [anchor=north west][inner sep=0.75pt]   [align=left] {{\small Motion planner }};
\draw (446,106) node [anchor=north west][inner sep=0.75pt]   [align=left] {{\small initialization}};
\draw (305,149) node [anchor=north west][inner sep=0.75pt]   [align=left] {{\small Scenario update}};
\draw (189,229) node [anchor=north west][inner sep=0.75pt]   [align=left] {{\small Offline policy evaluation}};
\draw (190,189) node [anchor=north west][inner sep=0.75pt]  [color=black  ,opacity=1 ] [align=left] {Pedestrian Simulator};
\draw (366,189) node [anchor=north west][inner sep=0.75pt]  [color=black  ,opacity=1 ] [align=left] {Pedestrian-aware Motion Planner};
\draw (191,269) node [anchor=north west][inner sep=0.75pt]   [align=left] {{\small Social force calculation}};
\draw (191,309) node [anchor=north west][inner sep=0.75pt]   [align=left] {{\small Pedestrian state update}};
\draw (397,229) node [anchor=north west][inner sep=0.75pt]   [align=left] {{\small Trajectory generation}};
\draw (397,269) node [anchor=north west][inner sep=0.75pt]   [align=left] {{\small Trajectory evaluation}};
\draw (403,309) node [anchor=north west][inner sep=0.75pt]   [align=left] {{\small Safety assessment}};
\draw (120,286) node [anchor=north west][inner sep=0.75pt]  [rotate=-270.0] [align=left] {{\small Next iteration}};

\end{tikzpicture}
}

%% file: Images/EvaluationFunnel.tex
\resizebox{0.55\linewidth}{!}{

\tikzset{every picture/.style={line width=1.0pt}} 

\begin{tikzpicture}[x=1.0pt,y=1.0pt,yscale=-1,xscale=1]

\draw  [draw opacity=0][fill={rgb, 255:red, 204; green, 204; blue, 204 }  ,fill opacity=1 ] (366,170) -- (377.23,200) -- (442.77,200) -- (454,170) -- cycle ;
\draw   (356,100) -- (366.2,130) -- (453.8,130) -- (464,100) -- cycle ;
\draw   (380,210) -- (440,210) -- (440,240) -- (380,240) -- cycle ;
\draw    (410,130) -- (410,137) ;
\draw [shift={(410,140)}, rotate = 270] [fill={rgb, 255:red, 0; green, 0; blue, 0 }  ][line width=0.08]  [draw opacity=0] (6.25,-3) -- (0,0) -- (6.25,3) -- cycle    ;
\draw   (366,170) -- (377.74,200) -- (442.26,200) -- (454,170) -- cycle ;
\draw   (366,140) -- (454,140) -- (454,160) -- (366,160) -- cycle ;
\draw    (410,160) -- (410,167) ;
\draw [shift={(410,170)}, rotate = 270] [fill={rgb, 255:red, 0; green, 0; blue, 0 }  ][line width=0.08]  [draw opacity=0] (6.25,-3) -- (0,0) -- (6.25,3) -- cycle    ;
\draw   (356,70) -- (464,70) -- (464,90) -- (356,90) -- cycle ;
\draw    (372,186) -- (340,186) -- (340,116) -- (359,116) ;
\draw [shift={(362,116)}, rotate = 180] [fill={rgb, 255:red, 0; green, 0; blue, 0 }  ][line width=0.08]  [draw opacity=0] (6.25,-3) -- (0,0) -- (6.25,3) -- cycle    ;
\draw    (410,200) -- (410,207) ;
\draw [shift={(410,210)}, rotate = 270] [fill={rgb, 255:red, 0; green, 0; blue, 0 }  ][line width=0.08]  [draw opacity=0] (6.25,-3) -- (0,0) -- (6.25,3) -- cycle    ;
\draw    (410,90) -- (410,97) ;
\draw [shift={(410,100)}, rotate = 270] [fill={rgb, 255:red, 0; green, 0; blue, 0 }  ][line width=0.08]  [draw opacity=0] (6.25,-3) -- (0,0) -- (6.25,3) -- cycle    ;

\draw (371,75) node [anchor=north west][inner sep=0.75pt][align=left] {{\small Trajectory generation}};
\draw (375,145) node [anchor=north west][inner sep=0.75pt][align=left] {{\small Selected trajectory}};
\draw (383,174) node [anchor=north west][inner sep=0.75pt][align=left] {{\small Harm and risk}};
\draw (369,115) node [anchor=north west][inner sep=0.75pt][align=left] {{\small Trajectory Evaluation}};
\draw (391,185) node [anchor=north west][inner sep=0.75pt][align=left] {{\small evaluation}};
\draw (387,104) node [anchor=north west][inner sep=0.75pt][align=left] {{\small Fundamental}};
\draw (386,215) node [anchor=north west][inner sep=0.75pt][align=left] {{\small Optimal and }};
\draw (383,226) node [anchor=north west][inner sep=0.75pt]   [align=left] {{\small safe trajectory}};
\draw (329,189) node [anchor=north west][inner sep=0.75pt]  [rotate=-270] [align=left] {{\small unsafe -- re-evaluate}};

\end{tikzpicture}
}

%% file: Images/simulation_step_duration.tex
\begin{tikzpicture}[font=\footnotesize]
  \begin{axis}[
    width=7.5cm,
    height=3.5cm,
    scale only axis,
    scaled ticks=false,
    scaled ticks=false,
    xmin=-3, xmax=73,
    boxplot/draw direction=x,
    xlabel={Calculation Time in \si{\milli\second}},
    ylabel={Number of Pedestrians},
    xtick={0, 10, ..., 70},
    ytick={1,2,3,4},
    yticklabels={61, 100, 208, 394},
    boxplot/box extend=0.8,
  ]
    \addplot+[
      mark=o,
      mark options={color=black},
      fill=TUMBlue,
      boxplot prepared={
        lower whisker=0.719308853149414,
        lower quartile=4.572510719299317,
        median=6.390810012817385,
        upper quartile=10.449051856994618,
        upper whisker=11.6579532623291,
        average=7.034791840447318
      },
      draw=black
    ] coordinates {
    };
    \addplot+[
      mark=o,
      mark options={color=black, fill=none},
      fill=Grey,
      boxplot prepared={
        lower whisker=1.56569480895996,
        lower quartile=6.1846375465393075,
        median=7.40182399749756,
        upper quartile=10.051131248474087,
        upper whisker=11.7354393005371,
        average=7.644348674350315
      },
      draw=black
    ] coordinates {
    };
    \addplot+[
      mark=o,
      mark options={color=black, fill=none, mark size=1.5pt},
      fill=TUMOrange,
      boxplot prepared={
        lower whisker=5.89680671691895,
        lower quartile=12.771785259246824,
        median=17.156004905700648,
        upper quartile=28.301000595092777,
        upper whisker=30.0338268280029,
        average=20.39179978547273
      },
      draw=black
    ] coordinates {
      (3, 76.9257545471191)
    };
    \addplot+[
      mark=o,
      mark options={color=black, fill=none, mark size=1.5pt},
      fill=TUMGreen,
      boxplot prepared={
        lower whisker=20.2629566192627,
        lower quartile=32.147109508514376,
        median=42.75429248809815,
        upper quartile=59.4940781593323,
        upper whisker=65.4222965240479,
        average=45.606277607105405
      },
      draw=black
    ] coordinates {
      (4, 108.146667480469)
    };
  \end{axis}
\end{tikzpicture}

%% file: Images/scenario2_risk.tex

\begin{axis}[
name=risk,
/pgf/number format/.cd,
1000 sep={},
height=3.3cm,
width=16cm,
legend style={
	at={(0.5,-0.25)}, 
	anchor=north,
	legend columns=5,
	cells={anchor=center},
	draw=none,
	column sep=0.25em,
	row sep=0.1em,
},
scale only axis,
scaled ticks=false,
scaled ticks=false,
tick label style={/pgf/number format/fixed},
ylabel={Risk $R$},
x label style={at={(0.5,-0.1)},anchor=north},
y label style={at={(-0.05,.5)},anchor=south},
xmin=0, xmax=140,
ymin=0, ymax=0.17,
ytick = {0.05, 0.10, 0.15},
yticklabels = {0.05, 0.10, 0.15},
ymajorgrids=true,
]

\addplot [ultra thick, TUMBlue, densely dotted]
table {%
1 0.00177225
2 0.00177225
4 0.005484
5 0.005484
7 0.0209488
8 0.0209488
10 0.0414073
11 0.0414073
13 0.0466321
14 0.0466321
16 0.049018
17 0.049018
19 0.0335923
20 0.0335923
22 0.0493847
23 0.0493847
25 0.0446448
26 0.0446448
28 0.0480959
29 0.0480959
31 0.0472354
32 0.0472354
34 0.0483508
35 0.0483508
37 0.0170894
38 0.0170894
40 0.010035
41 0.010035
43 0.0311124
44 0.0311124
46 0.0307826
47 0.0307826
49 0.00824661
50 0.00824661
52 0.00922131
53 0.00922131
55 0.00221998
56 0.00221998
58 0.0486567
59 0.0486567
61 0.00259897
62 0.00259897
64 0.0489779
65 0.0489779
67 0.0410285
68 0.0410285
70 0.0461324
71 0.0461324
73 0.0460681
74 0.0460681
76 0.0477694
77 0.0477694
79 0.00832484
80 0.00832484
82 2.11208e-05
83 2.11208e-05
85 4.44934e-05
86 4.44934e-05
88 0.000185724
89 0.000185724
91 0.000102251
92 0.000102251
94 0.00340303
95 0.00340303
97 0.0320331
98 0.0320331
100 0.00300422
101 0.00300422
103 0.0461658
104 0.0461658
106 0.0378607
107 0.0378607
109 0.049733
110 0.049733
112 0.040533
113 0.040533
115 0.0329512
116 0.0329512
118 0.00910005
119 0.00910005
121 0.00564953
122 0.00564953
124 0.0256366
125 0.0256366
127 0.0218688
128 0.0218688
130 0.0325889
131 0.0325889
133 0.0260459
134 0.0260459
136 0.0279807
};
\addlegendentry{$R_\mathrm{max} = 0.05$}

\addplot [ultra thick, TUMBlue, dashed]
table {%
1 0.00177225
2 0.00177225
4 0.00781255
5 0.00781255
7 0.0263145
8 0.0263145
10 0.0292416
11 0.0292416
13 0.0704332
14 0.0704332
16 0.0341263
17 0.0341263
19 0.0631551
20 0.0631551
22 0.0740494
23 0.0740494
25 0.0734256
26 0.0734256
28 0.0747774
29 0.0747774
31 0.0749468
32 0.0749468
34 0.0741513
35 0.0741513
37 0.0743916
38 0.0743916
40 0.022142
41 0.022142
43 0.0123419
44 0.0123419
46 0.00256745
47 0.00256745
49 0.001673
50 0.001673
52 0.0115955
53 0.0115955
55 0.00310899
56 0.00310899
58 0.0613316
59 0.0613316
61 0.074924
62 0.074924
64 0.0699609
65 0.0699609
67 0.0746043
68 0.0746043
70 0.064719
71 0.064719
73 0.0735874
74 0.0735874
76 0.0738957
77 0.0738957
79 0.0138308
80 0.0138308
82 1.77809e-05
83 1.77809e-05
85 7.80435e-05
86 7.80435e-05
88 0.000332618
89 0.000332618
91 0.000194856
92 0.000194856
94 0.000204812
95 0.000204812
97 0.00249026
98 0.00249026
100 0.00100103
101 0.00100103
103 0.0564526
104 0.0564526
106 0.0666063
107 0.0666063
109 0.0407321
110 0.0407321
112 0.0623177
113 0.0623177
115 0.0306696
116 0.0306696
118 0.010069
119 0.010069
121 0.0435373
122 0.0435373
124 0.00419461
125 0.00419461
127 0.0162024
128 0.0162024
130 0.0197946
131 0.0197946
133 0.0248703
134 0.0248703
136 0.0254958
137 0.0254958
139 0.0457008
140 0.0457008
142 0.0311932
143 0.0311932
145 0.0272886
146 0.0272886
148 0.0274341
149 0.0274341
151 0.0308691
152 0.0308691
154 0.0206049
155 0.0206049
157 0.034619
158 0.034619
160 0.0204144
161 0.0204144
163 0.0139868
164 0.0139868
166 0.0212569
167 0.0212569
169 0.0232828
170 0.0232828
172 0.0228546
};
\addlegendentry{$R_\mathrm{max} = 0.075$}

\addplot [ultra thick, TUMBlue, solid]
table {%
1 0.00177225
2 0.00177225
4 0.005484
5 0.005484
7 0.0209488
8 0.0209488
10 0.0414073
11 0.0414073
13 0.0959929
14 0.0959929
16 0.0704527
17 0.0704527
19 0.0853934
20 0.0853934
22 0.0992791
23 0.0992791
25 0.0894845
26 0.0894845
28 0.0999459
29 0.0999459
31 0.0998169
32 0.0998169
34 0.0949658
35 0.0949658
37 0.0686883
38 0.0686883
40 0.00152017
41 0.00152017
43 0.00504106
44 0.00504106
46 0.00193636
47 0.00193636
49 0.00101064
50 0.00101064
52 0.000964253
53 0.000964253
55 0.000782534
56 0.000782534
58 0.000430536
59 0.000430536
61 0.000893468
62 0.000893468
64 0.00114338
65 0.00114338
67 0.000991931
68 0.000991931
70 0.000819829
71 0.000819829
73 0.000653576
74 0.000653576
76 0.01079
77 0.01079
79 0.0947956
80 0.0947956
82 0.0458664
83 0.0458664
85 0.0785925
86 0.0785925
88 0.0558176
89 0.0558176
91 0.0827429
92 0.0827429
94 0.0996451
95 0.0996451
97 0.0426891
98 0.0426891
100 0.0108146
101 0.0108146
103 0.000124119
104 0.000124119
106 0.00350558
107 0.00350558
109 0.0167892
110 0.0167892
112 0.0215439
113 0.0215439
115 0.0167472
116 0.0167472
118 0.025676
119 0.025676
121 0.0510419
122 0.0510419
124 0.0220015
125 0.0220015
127 0.0611026
128 0.0611026
130 0.0183085
131 0.0183085
133 0.0417013
134 0.0417013
136 0.0297365
137 0.0297365
139 0.0299945
140 0.0299945
142 0.0436193
143 0.0436193
145 0.0145221
146 0.0145221
148 0.0163316
149 0.0163316
};
\addlegendentry{$R_\mathrm{max} = 0.10$}

\addplot [ultra thick, black]
table {%
1 0.00177225
2 0.00177225
4 0.005484
5 0.005484
7 0.0209488
8 0.0209488
10 0.0414073
11 0.0414073
13 0.125275
14 0.125275
16 0.118648
17 0.118648
19 0.114605
20 0.114605
22 0.111544
23 0.111544
25 0.107378
26 0.107378
28 0.10906
29 0.10906
31 0.0919169
32 0.0919169
34 0.0757322
35 0.0757322
37 0.0430763
38 0.0430763
40 0.000554729
41 0.000554729
43 0.00714778
44 0.00714778
46 0.00277637
47 0.00277637
49 0.000825027
50 0.000825027
52 0.000691472
53 0.000691472
55 0.000108964
56 0.000108964
58 8.71024e-05
59 8.71024e-05
61 0.00012411
62 0.00012411
64 0.00014226
65 0.00014226
67 0.000109493
68 0.000109493
70 8.11528e-05
71 8.11528e-05
73 0.000582118
74 0.000582118
76 0.0303605
77 0.0303605
79 0.134018
80 0.134018
82 0.162723
83 0.162723
85 0.104302
86 0.104302
88 0.0776209
89 0.0776209
91 0.060434
92 0.060434
94 0.14707
95 0.14707
97 0.00931537
98 0.00931537
100 0.0412611
101 0.0412611
103 0.0356921
104 0.0356921
106 0.0128337
107 0.0128337
109 0.0223794
110 0.0223794
112 0.0329211
113 0.0329211
115 0.0544241
116 0.0544241
118 0.158287
119 0.158287
121 0.0887747
122 0.0887747
124 0.0193512
125 0.0193512
127 0.0337423
128 0.0337423
130 0.152671
131 0.152671
133 0.0207793
134 0.0207793
136 0.0296176
137 0.0296176
139 0.045529
140 0.045529
142 0.0312927
143 0.0312927
145 0.0306288
};
\addlegendentry{$R_\mathrm{max} = \infty$}
\addplot [ultra thick, TUMOrange]
table {%
1 0.00177225
2 0.00177225
4 0.000792775
5 0.000792775
7 0.00115288
8 0.00115288
10 0.000390042
11 0.000390042
13 0.000367819
14 0.000367819
16 0.000526154
17 0.000526154
19 0.000287083
20 0.000287083
22 0.000337622
23 0.000337622
25 0.000831365
26 0.000831365
28 0.0001542
29 0.0001542
31 0.000270397
32 0.000270397
34 0.000370224
35 0.000370224
37 0.00043688
38 0.00043688
40 0.000481312
41 0.000481312
43 0.000525615
44 0.000525615
46 0.000807807
47 0.000807807
49 0.000939456
50 0.000939456
52 0.000995738
53 0.000995738
55 0.00117847
56 0.00117847
58 0.000323048
59 0.000323048
61 0.000429406
62 0.000429406
64 0.000367683
65 0.000367683
67 0.000543832
68 0.000543832
70 0.000100946
71 0.000100946
73 0.000209211
74 0.000209211
76 7.63413e-05
77 7.63413e-05
79 0.00120041
80 0.00120041
82 0.0136581
83 0.0136581
85 0.00981626
86 0.00981626
88 0.0158495
89 0.0158495
91 0.003089
92 0.003089
94 0.0332378
95 0.0332378
97 0.0190369
98 0.0190369
100 0.000613186
101 0.000613186
103 0.000191682
104 0.000191682
106 0.000130326
107 0.000130326
109 0.000100503
110 0.000100503
112 4.54609e-05
113 4.54609e-05
115 4.99724e-05
116 4.99724e-05
118 0.000227601
119 0.000227601
121 0.000220865
122 0.000220865
124 0.000251772
125 0.000251772
127 0.000269335
128 0.000269335
130 0.000341727
131 0.000341727
133 0.000401552
134 0.000401552
136 0.000362364
137 0.000362364
139 8.9118e-05
140 8.9118e-05
142 0.000167362
143 0.000167362
145 0.000179229
146 0.000179229
148 0.0002097
149 0.0002097
151 3.2976e-05
152 3.2976e-05
154 1.10728e-05
155 1.10728e-05
157 5.28924e-06
158 5.28924e-06
160 5.4841e-06
161 5.4841e-06
163 5.5233e-06
164 5.5233e-06
166 2.20492e-05
167 2.20492e-05
169 4.21715e-05
170 4.21715e-05
172 0.000108056
173 0.000108056
175 1.32229e-05
176 1.32229e-05
178 1.34639e-05
179 1.34639e-05
181 0.000414513
182 0.000414513
184 0.000374182
185 0.000374182
187 0.000425507
188 0.000425507
190 0.000551391
191 0.000551391
199 0
200 0
262 0
};
\addlegendentry{Collision prob}

\addplot[draw=black, line width=1pt, dashed] coordinates {(30,0) (30,0.2)};

\addplot[draw=black, line width=1pt, dashed] coordinates {(90,0) (90,0.2)};

\legend{};

\end{axis}


%% file: Images/scenario2_velocity.tex
\begin{tikzpicture}[font=\footnotesize, trim axis right]

\begin{axis}[
/pgf/number format/.cd,
1000 sep={},
height=3.3cm,
width=16cm,
legend style={
	at={(0.5,-0.2)}, 
	anchor=north,
	legend columns=5,
	cells={anchor=center},
	draw=none,
	column sep=0.25em,
	row sep=0.1em,
},
scale only axis,
scaled ticks=false,
scaled ticks=false,
tick label style={/pgf/number format/fixed},
xlabel={Timestep $t$},
ylabel={Velocity $v$ in \si{\meter\per\second}},
x label style={at={(0.5,-0.1)},anchor=north},
y label style={at={(-0.05,.5)},anchor=south},
xmin=0, xmax=140,
ymin=0, ymax=6.4,
ytick = {1, 2, ..., 6},
ymajorgrids=true
]

\addplot [ultra thick, TUMBlue, densely dotted]
table {%
1 4.00704
2 4.02735
4 4.0847
5 4.11994
7 4.19434
8 4.2329
10 4.31034
11 4.34942
13 4.43446
14 4.49413
16 4.60278
17 4.60363
19 4.55052
20 4.50414
22 4.39376
23 4.32578
25 4.16621
26 4.04919
28 3.79106
29 3.65738
31 3.38994
32 3.24256
34 2.96676
35 2.87033
37 2.69238
38 2.55008
40 2.33279
41 2.45115
43 2.85886
44 3.09566
46 3.55479
47 3.77994
49 4.16844
50 4.28594
52 4.44857
53 4.51676
55 4.65361
56 4.75298
58 4.95889
59 5.03774
61 5.16718
62 5.22194
64 5.3011
65 5.28812
67 5.20078
68 5.11879
70 4.91034
71 4.76321
73 4.40616
74 4.14216
76 3.61602
77 3.42473
79 3.20458
80 3.30664
82 3.66433
83 3.87282
85 4.27735
86 4.47575
88 4.82184
89 4.93578
91 5.0977
92 5.1449
94 5.21177
95 5.24118
97 5.29521
98 5.32074
100 5.36821
101 5.391
103 5.42272
104 5.4098
106 5.3559
107 5.31807
109 5.23904
110 5.20755
112 5.1515
113 5.11704
115 5.05596
116 5.05733
118 5.08415
119 5.10536
121 5.15346
122 5.18134
124 5.23182
125 5.24544
127 5.2603
128 5.26297
130 5.26425
131 5.26411
133 5.26402
134 5.26424
136 5.2656
};
\addlegendentry{$R_\mathrm{max} = 0.05$}

\addplot [ultra thick, TUMBlue, dashed]
table {%
1 4.00307
2 4.0118
4 4.03631
5 4.05122
7 4.08238
8 4.09814
10 4.12969
11 4.14546
13 4.17269
14 4.17615
16 4.16937
17 4.15244
19 4.10633
20 4.07831
22 4.01568
23 3.97548
25 3.86429
26 3.75332
28 3.52021
29 3.43953
31 3.32056
32 3.27834
34 3.21005
35 3.18251
37 3.13397
38 3.11225
40 3.11106
41 3.20236
43 3.44898
44 3.5697
46 3.78772
47 3.8892
49 4.07435
50 4.16098
52 4.31966
53 4.39409
55 4.52628
56 4.57955
58 4.66207
59 4.67891
61 4.73214
62 4.85039
64 5.09831
65 5.12069
67 5.06136
68 4.99648
70 4.84781
71 4.77319
73 4.6282
74 4.55739
76 4.41917
77 4.35023
79 4.2412
80 4.25032
82 4.33066
83 4.39129
85 4.51247
86 4.56365
88 4.64847
89 4.6772
91 4.7207
92 4.73697
94 4.76433
95 4.77598
97 4.79703
98 4.80679
100 4.82518
101 4.83388
103 4.8486
104 4.85207
106 4.85471
107 4.8546
109 4.84611
110 4.82408
112 4.77055
113 4.75687
115 4.74535
116 4.7457
118 4.75389
119 4.76228
121 4.78232
122 4.79381
124 4.81526
125 4.8227
127 4.83492
128 4.83942
130 4.84617
131 4.84849
133 4.85154
134 4.85243
136 4.8536
137 4.85402
139 4.85479
140 4.85512
142 4.85554
143 4.85564
145 4.85542
146 4.85517
148 4.85462
149 4.85433
151 4.85384
152 4.85368
154 4.85343
155 4.85334
157 4.85318
158 4.85312
160 4.85306
161 4.85307
163 4.85313
164 4.85317
166 4.85322
167 4.85322
169 4.85312
170 4.85304
172 4.85283
};
\addlegendentry{$R_\mathrm{max} = 0.075$}

\addplot [ultra thick, TUMBlue, solid]
table {%
1 4.00704
2 4.02735
4 4.0847
5 4.11994
7 4.19434
8 4.2329
10 4.31034
11 4.34942
13 4.43837
14 4.50914
16 4.6503
17 4.67586
19 4.71012
20 4.78545
22 4.96116
23 5.00681
25 5.04356
26 5.03153
28 4.98265
29 4.95385
31 4.90043
32 4.88482
34 4.87677
35 4.90165
37 4.96943
38 5.00476
40 5.07531
41 5.11057
43 5.17726
44 5.20933
46 5.26853
47 5.29573
49 5.34349
50 5.36479
52 5.4008
53 5.41722
55 5.4475
56 5.46218
58 5.49064
59 5.50444
61 5.52985
62 5.54149
64 5.56211
65 5.57141
67 5.58781
68 5.59537
70 5.6094
71 5.61629
73 5.62912
74 5.63517
76 5.64571
77 5.65019
79 5.65592
80 5.65765
82 5.65479
83 5.64025
85 5.59658
86 5.56881
88 5.50382
89 5.45919
91 5.36419
92 5.32306
94 5.24817
95 5.21263
97 5.14714
98 5.13007
100 5.12088
101 5.1257
103 5.14851
104 5.17169
106 5.22284
107 5.24153
109 5.26861
110 5.27716
112 5.28545
113 5.27813
115 5.2514
116 5.23322
118 5.19295
119 5.17162
121 5.13351
122 5.12513
124 5.11913
125 5.1188
127 5.11953
128 5.12006
130 5.12093
131 5.12123
133 5.11826
134 5.10853
136 5.08508
137 5.08035
139 5.07751
140 5.07714
142 5.07679
143 5.0767
145 5.07668
146 5.07678
148 5.07713
149 5.07735
};
\addlegendentry{$R_\mathrm{max} = 0.10$}

\addplot [ultra thick, black]
table {%
1 4.00704
2 4.02735
4 4.0847
5 4.11994
7 4.19434
8 4.2329
10 4.31034
11 4.34942
13 4.42694
14 4.46544
16 4.54213
17 4.581
19 4.65771
20 4.69599
22 4.77101
23 4.80814
25 4.88076
26 4.91672
28 4.98741
29 5.02264
31 5.09202
32 5.1266
34 5.19465
35 5.22852
37 5.29175
38 5.31494
40 5.35293
41 5.37204
43 5.40837
44 5.42568
46 5.45746
47 5.47243
49 5.49997
50 5.51297
52 5.5368
53 5.54798
55 5.56831
56 5.57776
58 5.59493
59 5.60295
61 5.6176
62 5.62446
64 5.63708
65 5.643
67 5.65384
68 5.6589
70 5.66812
71 5.67243
73 5.68026
74 5.68392
76 5.6907
77 5.69396
79 5.7
80 5.70288
82 5.70825
83 5.71084
85 5.71587
86 5.71835
88 5.72297
89 5.72517
91 5.72931
92 5.73136
94 5.73548
95 5.73757
97 5.64048
98 5.36594
100 4.77982
101 4.7093
103 4.78142
104 4.87398
106 5.07051
107 5.1432
109 5.2545
110 5.29068
112 5.34431
113 5.36657
115 5.40596
116 5.42411
118 5.45707
119 5.47242
121 5.5003
122 5.51329
124 5.53688
125 5.5479
127 5.56801
128 5.57744
130 5.59468
131 5.60275
133 5.61748
134 5.62439
136 5.63701
137 5.64294
139 5.65376
140 5.65883
142 5.66803
143 5.6723
145 5.68008
};
\addlegendentry{$R_\mathrm{max} = \infty$}
\addplot [ultra thick, TUMOrange]
table {%
1 4.04886
2 4.18307
4 4.49307
5 4.59169
7 4.71543
8 4.73763
10 4.73777
11 4.70445
13 4.60169
14 4.52505
16 4.32705
17 4.16272
19 3.78082
20 3.54594
22 3.08054
23 2.86636
25 2.48568
26 2.32146
28 2.01367
29 1.82902
31 1.46351
32 1.29879
34 1.0116
35 0.891815
37 0.689147
38 0.606416
40 0.467595
41 0.411265
43 0.316976
44 0.278785
46 0.214902
47 0.18904
49 0.145789
50 0.128282
52 0.0990056
53 0.0871555
55 0.0673391
56 0.0593182
58 0.0630289
59 0.107303
61 0.234614
62 0.305732
64 0.455734
65 0.546395
67 0.77447
68 0.970272
70 1.37234
71 1.51522
73 1.73065
74 1.80726
76 1.9206
77 1.94043
79 1.94475
80 1.93582
82 1.9247
83 1.94803
85 2.02278
86 2.06744
88 2.14865
89 2.19661
91 2.28984
92 2.32542
94 2.39216
95 2.43918
97 2.54476
98 2.61695
100 2.76239
101 2.79229
103 2.80261
104 2.78784
106 2.74123
107 2.71026
109 2.64032
110 2.59634
112 2.4905
113 2.40936
115 2.22415
116 2.11128
118 1.90152
119 1.83444
121 1.74232
122 1.71187
124 1.6639
125 1.64354
127 1.60443
128 1.58548
130 1.54764
131 1.52931
133 1.49304
134 1.47569
136 1.44171
137 1.42575
139 1.39449
140 1.37994
142 1.35269
143 1.34036
145 1.31796
146 1.30386
148 1.27519
149 1.25902
151 1.22516
152 1.20192
154 1.15133
155 1.12652
157 1.0787
158 1.05557
160 1.01082
161 0.989101
163 0.947113
164 0.926771
166 0.890152
167 0.878983
169 0.85844
170 0.83706
172 0.78681
173 0.764835
175 0.731389
176 0.729401
178 0.737409
179 0.743957
181 0.75933
182 0.771734
184 0.800955
185 0.816978
187 0.849672
188 0.866316
190 0.899353
191 0.915798
193 0.945912
194 0.955509
196 0.968679
197 0.972373
199 0.97707
200 0.978251
202 0.979347
203 0.979331
205 0.978832
206 0.978998
208 0.979608
209 0.979948
211 0.980516
212 0.980817
214 0.983055
215 0.98848
217 1.00116
218 1.00288
220 1.00484
221 1.01081
223 1.11024
224 1.34981
226 1.9749
227 2.32875
229 2.99099
230 3.27001
232 3.71808
233 3.85285
235 4.03028
236 4.07795
238 4.13298
239 4.13838
241 4.1278
242 4.11001
244 4.06416
245 4.03719
247 3.97657
248 3.93448
250 3.8375
251 3.78015
253 3.6572
254 3.58777
256 3.43623
257 3.33905
259 3.1253
260 2.99946
262 2.73343
};
\addlegendentry{Baseline Planner}

\addplot[draw=black, line width=1pt, dashed] coordinates {(30,0) (30,7)};

\addplot[draw=black, line width=1pt, dashed] coordinates {(90,0) (90,7)};

\end{axis}

\end{tikzpicture}

%% file: Images/scenario3_crosswalk_velocity_trajectory_number.tex
\newcommand{\plotheight}{2.8cm}

\begin{tikzpicture}[font=\footnotesize]


\node[inner sep=0pt, anchor=south west, draw=none, dashed, line width=1pt] (img1) at (0, \plotheight+0.05) 
{\includegraphics[width=7cm, trim={0.0cm 0.0cm 0cm 0.0cm},clip]{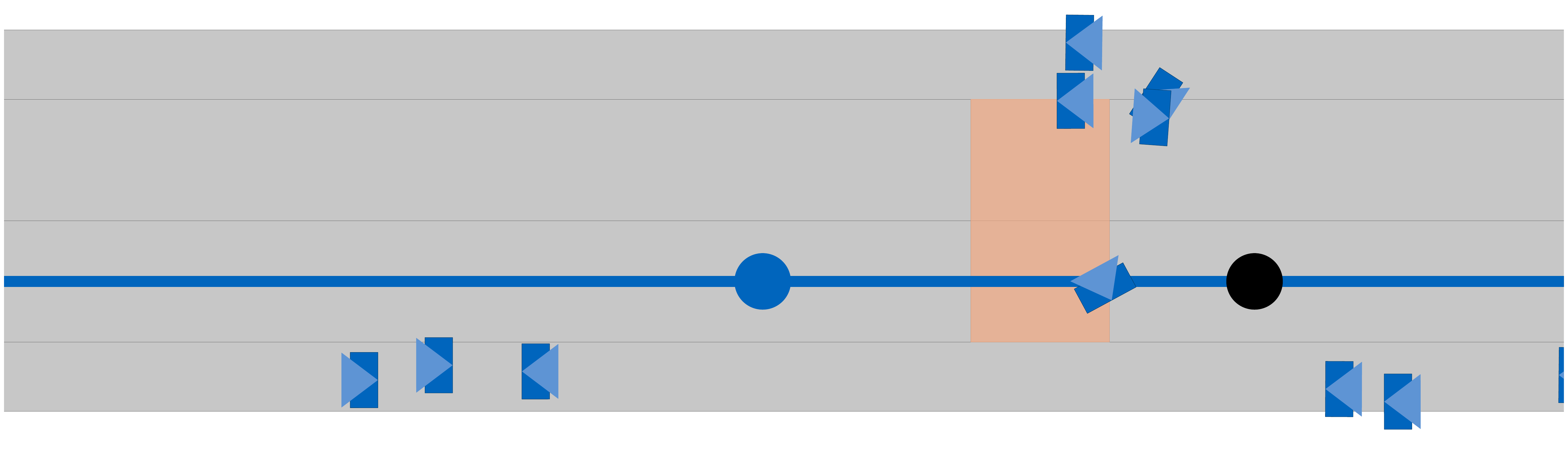}};


\begin{axis}[
/pgf/number format/.cd,
1000 sep={},
height=\plotheight,
width=7cm,
scale only axis,
scaled ticks=false,
scaled ticks=false,
tick label style={/pgf/number format/fixed},
xlabel={Coordinate $s$ in \si{\meter}},
ylabel={Timestep $t$},
xmin=70, xmax=115,
ymin=0, ymax=190,
axis y line*=left,
]

\addplot[forget plot, draw=none, fill=policy-crosswalk, fill opacity=0.7, line width=5pt] coordinates {(97.89,-1) (97.89,200) (101.89,200) (101.89,-1)};

\addplot [ultra thick, TUMBlue]
table {%
59.0776 1
59.5776 2
60.5776 4
61.0776 5
62.0776 7
62.5776 8
63.5776 10
64.0776 11
65.0776 13
65.5776 14
66.5776 16
67.0776 17
68.0776 19
68.5776 20
69.5776 22
70.0776 23
71.0776 25
71.5776 26
72.5776 28
73.0776 29
74.0774 31
74.5761 32
75.5667 34
76.0582 35
77.0306 37
77.5119 38
78.462 40
78.9313 41
79.8561 43
80.3121 44
81.2085 46
81.6497 47
82.5147 49
82.9394 50
83.7694 52
84.1756 53
84.9664 55
85.3519 56
86.0986 58
86.4609 59
87.1581 61
87.4941 62
88.1352 64
88.4416 65
89.0194 67
89.2924 68
89.7987 70
90.034 71
90.4605 73
90.654 74
90.9927 76
91.141 77
91.3871 79
91.4894 80
91.6455 82
91.7055 83
91.7871 85
91.8158 86
91.8508 88
91.8623 89
91.876 91
91.8803 92
91.8855 94
91.8871 95
91.8892 97
91.8899 98
91.8907 100
91.8909 101
91.8912 103
91.8913 104
91.8915 106
91.8915 107
91.8915 109
91.8916 110
91.8916 112
91.8916 113
91.8916 115
91.8916 116
91.8916 118
91.8916 119
91.8956 121
91.9218 122
92.0956 124
92.2389 125
92.6477 127
92.8983 128
93.4878 130
93.8124 131
94.5096 133
94.8756 134
95.6422 136
96.0392 137
96.8621 139
97.2852 140
98.1558 142
98.601 143
99.5122 145
99.9747 146
100.91 148
101.381 149
102.328 151
102.804 152
103.76 154
104.239 155
105.201 157
105.684 158
106.649 160
107.133 161
108.099 163
108.583 164
109.55 166
110.034 167
111.001 169
111.485 170
112.452 172
112.936 173
113.903 175
114.387 176
115.354 178
115.838 179
};
\label{plot:0075-trajnumber-solid}

\addplot [ultra thick, black]
table {%
59.0776 1
59.5776 2
60.5776 4
61.0776 5
62.0776 7
62.5776 8
63.5776 10
64.0776 11
65.0776 13
65.5776 14
66.5776 16
67.0776 17
68.0776 19
68.5776 20
69.5776 22
70.0776 23
71.0776 25
71.5776 26
72.5776 28
73.0776 29
74.0776 31
74.5776 32
75.5776 34
76.0776 35
77.0776 37
77.5776 38
78.5776 40
79.0776 41
80.0776 43
80.5776 44
81.5776 46
82.0776 47
83.0776 49
83.5776 50
84.5776 52
85.0776 53
86.0776 55
86.5776 56
87.5776 58
88.0776 59
89.0776 61
89.5776 62
90.5776 64
91.0776 65
92.0776 67
92.5776 68
93.5776 70
94.0776 71
95.0776 73
95.5776 74
96.5776 76
97.0776 77
98.0776 79
98.5776 80
99.5776 82
100.078 83
101.078 85
101.578 86
102.578 88
103.078 89
104.078 91
104.578 92
105.578 94
106.078 95
107.078 97
107.578 98
108.578 100
109.078 101
110.078 103
110.578 104
111.578 106
112.078 107
113.078 109
113.578 110
114.578 112
115.078 113
116.078 115
116.578 116
117.578 118
118.078 119
119.078 121
119.578 122
120.578 124
121.078 125
122.078 127
122.578 128
123.578 130
124.078 131
125.078 133
125.578 134
126.578 136
127.078 137
};
\label{plot:aggressive-trajnumber-solid}

\addplot [only marks, mark=*, mark size=3pt, mark options={color=TUMBlue}] coordinates {(91.89, 95)};

\addplot [only marks, mark=*, mark size=3pt, mark options={color=black}] coordinates {(106.077, 95)};

\end{axis}


\begin{axis}[
/pgf/number format/.cd,
1000 sep={},
height=\plotheight,
width=7cm,
legend style={
	at={(0.5,-0.25)}, 
	anchor=north,
	legend columns=2,
	cells={anchor=center},
	draw=none,
	column sep=0.25em,
	row sep=0.1em,
},
scale only axis,
scaled ticks=false,
scaled ticks=false,
tick label style={/pgf/number format/fixed},
ylabel={Velocity $v$ in \si{\meter\per\second}},
xmin=70, xmax=115,
ymin=0, 
ymax=5.5,
axis x line*=none,
axis y line*=right,
xtick = {-10},
ytick = {0, 2, 4}
]

\addplot [ultra thick, TUMBlue, dashed]
table {%
59.0776 5
59.5776 5
60.5776 5
61.0776 5
62.0776 5
62.5776 5
63.5776 5
64.0776 5
65.0776 5
65.5776 5
66.5776 5
67.0776 5
68.0776 5
68.5776 5
69.5776 5
70.0776 5
71.0776 5
71.5776 5
72.5776 5
73.0776 5
74.0774 4.99437
74.5761 4.97778
75.5667 4.92956
76.0582 4.89868
77.0306 4.83071
77.5119 4.79303
78.462 4.71391
78.9313 4.67132
79.8561 4.58278
80.3121 4.53538
81.2085 4.43697
81.6497 4.38425
82.5147 4.27472
82.9394 4.21593
83.7694 4.09363
84.1756 4.02783
84.9664 3.89077
85.3519 3.81686
86.0986 3.66271
86.4609 3.57944
87.1581 3.40555
87.4941 3.31151
88.1352 3.11497
88.4416 3.00873
89.0194 2.78663
89.2924 2.66691
89.7987 2.41694
90.034 2.28321
90.4605 2.00503
90.654 1.85865
90.9927 1.55679
91.141 1.40329
91.3871 1.09247
91.4894 0.945201
91.6455 0.656994
91.7055 0.539211
91.7871 0.318974
91.8158 0.251983
91.8508 0.126922
91.8623 0.10145
91.876 0.0475621
91.8803 0.0372284
91.8855 0.0187387
91.8871 0.0147378
91.8892 0.00733829
91.8899 0.00576936
91.8907 0.00287631
91.8909 0.00226113
91.8912 0.00112702
91.8913 0.000886103
91.8915 0.000441717
91.8915 0.000347243
91.8915 0.00017308
91.8916 0.000136081
91.8916 6.78351e-05
91.8916 5.33268e-05
91.8916 2.41738e-05
91.8916 1.5854e-05
91.8916 0
91.8916 0
91.8956 0.117184
91.9218 0.438496
92.0956 1.23861
92.2389 1.63882
92.6477 2.35872
92.8983 2.66045
93.4878 3.15839
93.8124 3.33256
94.5096 3.60305
94.8756 3.71872
95.6422 3.92415
96.0392 4.01896
96.8621 4.19136
97.2852 4.27188
98.1558 4.41878
98.601 4.4875
99.5122 4.60547
99.9747 4.64396
100.91 4.69969
101.381 4.71942
102.328 4.75156
102.804 4.76486
103.76 4.78875
104.239 4.79971
105.201 4.81893
105.684 4.82446
106.649 4.83119
107.133 4.83293
108.099 4.83505
108.583 4.8356
109.55 4.83626
110.034 4.83644
111.001 4.83665
111.485 4.8367
112.452 4.83677
112.936 4.83678
113.903 4.83681
114.387 4.83681
115.354 4.83682
115.838 4.83682
};
\label{plot:0075-velocity-dashed}

\addplot [ultra thick, black, dashed]
table {%
59.0776 5
59.5776 5
60.5776 5
61.0776 5
62.0776 5
62.5776 5
63.5776 5
64.0776 5
65.0776 5
65.5776 5
66.5776 5
67.0776 5
68.0776 5
68.5776 5
69.5776 5
70.0776 5
71.0776 5
71.5776 5
72.5776 5
73.0776 5
74.0776 5
74.5776 5
75.5776 5
76.0776 5
77.0776 5
77.5776 5
78.5776 5
79.0776 5
80.0776 5
80.5776 5
81.5776 5
82.0776 5
83.0776 5
83.5776 5
84.5776 5
85.0776 5
86.0776 5
86.5776 5
87.5776 5
88.0776 5
89.0776 5
89.5776 5
90.5776 5
91.0776 5
92.0776 5
92.5776 5
93.5776 5
94.0776 5
95.0776 5
95.5776 5
96.5776 5
97.0776 5
98.0776 5
98.5776 5
99.5776 5
100.078 5
101.078 5
101.578 5
102.578 5
103.078 5
104.078 5
104.578 5
105.578 5
106.078 5
107.078 5
107.578 5
108.578 5
109.078 5
110.078 5
110.578 5
111.578 5
112.078 5
113.078 5
113.578 5
114.578 5
115.078 5
116.078 5
116.578 5
117.578 5
118.078 5
119.078 5
119.578 5
120.578 5
121.078 5
122.078 5
122.578 5
123.578 5
124.078 5
125.078 5
125.578 5
126.578 5
127.078 5
};
\label{plot:aggressive-velocity-dashed}

\addplot [only marks, mark=halfcircle, mark size=3pt, mark color=TUMBlue, mark options={color=TUMBlue}] coordinates {(91.89, 0.015)};

\addplot [only marks, mark=halfcircle, mark size=3pt, mark color=black, mark options={color=black}] coordinates {(106.077, 4.999)};

\legend{}

\end{axis}


\begin{axis}[
    height=\plotheight,  
    width=7cm,
    hide axis,
    axis x line=none,
    axis y line=none,
    tick style=none,
    enlarge x limits=false,
    enlarge y limits=false,
    xmin=0, xmax=1,
    ymin=0, ymax=1,
    legend columns=2,
    legend style=
    {
        at={(0.65,-0.65)}, 
        anchor=north,
        legend columns=2,
        cells={anchor=center},
        draw=none,
        column sep=0.25em,
        row sep=0.1em,
    },
]
\addlegendimage{ultra thick, TUMBlue}
\addlegendentry{$R_\mathrm{max}=0.075$ (Timestep)}

\addlegendimage{ultra thick, TUMBlue, dashed}
\addlegendentry{$R_\mathrm{max}=0.075$ (Velocity)}

\addlegendimage{ultra thick, black}
\addlegendentry{$R_\mathrm{max}=\infty$ (Timestep)}

\addlegendimage{ultra thick, black, dashed}
\addlegendentry{$R_\mathrm{max}=\infty$ (Velocity)}

\end{axis}

\end{tikzpicture}

%% file: main.bbl
\begin{thebibliography}{10}
\providecommand{\url}[1]{#1}
\csname url@samestyle\endcsname
\providecommand{\newblock}{\relax}
\providecommand{\bibinfo}[2]{#2}
\providecommand{\BIBentrySTDinterwordspacing}{\spaceskip=0pt\relax}
\providecommand{\BIBentryALTinterwordstretchfactor}{4}
\providecommand{\BIBentryALTinterwordspacing}{\spaceskip=\fontdimen2\font plus
\BIBentryALTinterwordstretchfactor\fontdimen3\font minus \fontdimen4\font\relax}
\providecommand{\BIBforeignlanguage}[2]{{%
\expandafter\ifx\csname l@#1\endcsname\relax
\typeout{** WARNING: IEEEtran.bst: No hyphenation pattern has been}%
\typeout{** loaded for the language `#1'. Using the pattern for}%
\typeout{** the default language instead.}%
\else
\language=\csname l@#1\endcsname
\fi
#2}}
\providecommand{\BIBdecl}{\relax}
\BIBdecl

\bibitem{Bobisse2019}
R.~Bobisse and A.~Pavia, \emph{Automatic for the City: Designing for People in the Age of the Driverless Car}.\hskip 1em plus 0.5em minus 0.4em\relax RIBA Publishing, Aug. 2019.

\bibitem{pendleton2017perception}
S.~D. Pendleton, H.~Andersen, X.~Du, X.~Shen, M.~Meghjani, Y.~H. Eng, D.~Rus, and M.~H. Ang, ``Perception, planning, control, and coordination for autonomous vehicles,'' \emph{Machines}, vol.~5, no.~1, p.~6, 2017.

\bibitem{Pokorny2022}
P.~Pokorny and A.~Høye, ``Descriptive analysis of reports on autonomous vehicle collisions in california: January 2021–june 2022,'' \emph{Traffic Safety Research}, vol.~2, Sep. 2022.

\bibitem{Taxonomy2024}
J.~Betz, M.~Lutwitzi, and S.~Peters, ``A new taxonomy for automated driving: Structuring applications based on their operational design domain, level of automation and automation readiness,'' in \emph{2024 IEEE Intelligent Vehicles Symposium (IV)}, 2024, pp. 1--7.

\bibitem{Huang.2023}
J.~Huang, A.~Gautam, and S.~Saripalli, ``Learning pedestrian actions to ensure safe autonomous driving,'' in \emph{2023 IEEE Intelligent Vehicles Symposium (IV)}.\hskip 1em plus 0.5em minus 0.4em\relax IEEE, 2023.

\bibitem{rasouli_autonomous_2020}
A.~Rasouli and J.~K. Tsotsos, ``\BIBforeignlanguage{en}{Autonomous {Vehicles} {That} {Interact} {With} {Pedestrians}: {A} {Survey} of {Theory} and {Practice}},'' \emph{\BIBforeignlanguage{en}{IEEE Transactions on Intelligent Transportation Systems}}, vol.~21, no.~3, 2020.

\bibitem{Dosovitskiy17}
A.~Dosovitskiy, G.~Ros, F.~Codevilla, A.~Lopez, and V.~Koltun, ``{CARLA}: {An} open urban driving simulator,'' in \emph{Proceedings of the 1st Annual Conference on Robot Learning}, 2017, pp. 1--16.

\bibitem{althoff_commonroad_2017}
M.~Althoff, M.~Koschi, and S.~Manzinger, ``{CommonRoad}: {Composable} benchmarks for motion planning on roads,'' in \emph{2017 {IEEE} {Intelligent} {Vehicles} {Symposium} ({IV})}, Jun. 2017.

\bibitem{UN_Regulation_157}
{United Nations}, ``United nations regulation no. 157: Uniform provisions concerning the approval of vehicles with regard to automated lane keeping systems,'' Geneva, 2020.

\bibitem{ciuffo_2024}
B.~Ciuffo, R.~Donà, M.~Galassi, W.~Giannotti, C.~Sollima, F.~Terzuoli, and S.~Vass, ``Interpretation of eu regulation 2022/1426 on the type approval of automated driving systems,'' Luxembourg, 2024.

\bibitem{karle2022scenario}
P.~Karle, M.~Geisslinger, J.~Betz, and M.~Lienkamp, ``Scenario understanding and motion prediction for autonomous vehicles—review and comparison,'' \emph{IEEE Transactions on Intelligent Transportation Systems}, vol.~23, no.~10, pp. 16\,962--16\,982, 2022.

\bibitem{EU_Regulation_2019_2144}
{European Union}, ``Regulation (eu) 2019/2144 of the european parliament and of the council of 27 november 2019 on type-approval requirements for motor vehicles and their trailers, and systems, components and separate technical units intended for such vehicles, as regards their general safety and the protection of vehicle occupants and vulnerable road users,'' 2019.

\bibitem{haus_estimated_2019}
S.~H. Haus, R.~Sherony, and H.~C. Gabler, ``Estimated benefit of automated emergency braking systems for vehicle–pedestrian crashes in the {United} {States},'' \emph{Traffic Injury Prevention}, vol.~20, 2019.

\bibitem{CICCHINO2022106686}
J.~B. Cicchino, ``Effects of automatic emergency braking systems on pedestrian crash risk,'' \emph{Accident Analysis \& Prevention}, vol. 172, 2022.

\bibitem{sprenger_cross-cultural_2023}
J.~Sprenger, L.~Hell, M.~Klusch, Y.~Kobayashi, S.~Kudo, and C.~Müller, ``\BIBforeignlanguage{en}{Cross-{Cultural} {Behavior} {Analysis} of {Street}-{Crossing} {Pedestrians} in {Japan} and {Germany}},'' in \emph{\BIBforeignlanguage{en}{2023 {IEEE} {Intelligent} {Vehicles} {Symposium} ({IV})}}, Anchorage, AK, USA, 2023.

\bibitem{camara_pedestrian_2021-1}
F.~Camara, N.~Bellotto, S.~Cosar, D.~Nathanael, M.~Althoff, J.~Wu, J.~Ruenz, A.~Dietrich, and C.~W. Fox, ``\BIBforeignlanguage{en}{Pedestrian {Models} for {Autonomous} {Driving} {Part} {I}: {Low}-{Level} {Models}, {From} {Sensing} to {Tracking}},'' \emph{\BIBforeignlanguage{en}{IEEE Transactions on Intelligent Transportation Systems}}, vol.~22, no.~10, 2021.

\bibitem{camara_pedestrian_2021-2}
F.~Camara, N.~Bellotto, S.~Cosar, F.~Weber, D.~Nathanael, M.~Althoff, J.~Wu, J.~Ruenz, A.~Dietrich, G.~Markkula, A.~Schieben, F.~Tango, N.~Merat, and C.~Fox, ``\BIBforeignlanguage{en}{Pedestrian {Models} for {Autonomous} {Driving} {Part} {II}: {High}-{Level} {Models} of {Human} {Behavior}},'' \emph{\BIBforeignlanguage{en}{IEEE Transactions on Intelligent Transportation Systems}}, vol.~22, no.~9, 2021.

\bibitem{rudenko_human_2020}
A.~Rudenko, L.~Palmieri, M.~Herman, K.~M. Kitani, D.~M. Gavrila, and K.~O. Arras, ``\BIBforeignlanguage{en}{Human motion trajectory prediction: a survey},'' \emph{\BIBforeignlanguage{en}{The International Journal of Robotics Research}}, vol.~39, no.~8, Jul. 2020.

\bibitem{gulzar_survey_2021}
M.~Gulzar, Y.~Muhammad, and N.~Muhammad, ``A {Survey} on {Motion} {Prediction} of {Pedestrians} and {Vehicles} for {Autonomous} {Driving},'' \emph{IEEE Access}, vol.~9, 2021.

\bibitem{sharma_pedestrian_2022}
N.~Sharma, C.~Dhiman, and S.~Indu, ``Pedestrian {Intention} {Prediction} for {Autonomous} {Vehicles}: {A} {Comprehensive} {Survey},'' \emph{Neurocomputing}, vol. 508, 2022.

\bibitem{zhang_pedestrian_2023}
C.~Zhang and C.~Berger, ``Pedestrian {Behavior} {Prediction} {Using} {Deep} {Learning} {Methods} for {Urban} {Scenarios}: {A} {Review},'' \emph{IEEE Transactions on Intelligent Transportation Systems}, vol.~24, no.~10, Oct. 2023.

\bibitem{korbmacher_review_2022}
R.~Korbmacher and A.~Tordeux, ``Review of {Pedestrian} {Trajectory} {Prediction} {Methods}: {Comparing} {Deep} {Learning} and {Knowledge}-{Based} {Approaches},'' \emph{IEEE Transactions on Intelligent Transportation Systems}, vol.~23, no.~12, Dec. 2022.

\bibitem{korbmacher_deep_2024}
------, ``\BIBforeignlanguage{en}{Deep {Learning} for {Predicting} {Pedestrian} {Trajectories} in {Crowds}},'' in \emph{\BIBforeignlanguage{en}{Intelligent {Systems} and {Applications}}}, K.~Arai, Ed.\hskip 1em plus 0.5em minus 0.4em\relax Cham: Springer Nature Switzerland, 2024.

\bibitem{kalatian2022context}
A.~Kalatian and B.~Farooq, ``A context-aware pedestrian trajectory prediction framework for automated vehicles,'' \emph{Transportation research part C: emerging technologies}, vol. 134, p. 103453, 2022.

\bibitem{azarmi2023local}
M.~Azarmi, M.~Rezaei, T.~Hussain, and C.~Qian, ``Local and global contextual features fusion for pedestrian intention prediction,'' in \emph{International Conference on Artificial Intelligence and Smart Vehicles}.\hskip 1em plus 0.5em minus 0.4em\relax Springer, 2023, pp. 1--13.

\bibitem{fang2024behavioral}
J.~Fang, F.~Wang, J.~Xue, and T.-S. Chua, ``Behavioral intention prediction in driving scenes: A survey,'' \emph{IEEE Transactions on Intelligent Transportation Systems}, 2024.

\bibitem{Paden2016}
B.~Paden, M.~Čáp, S.~Z. Yong, D.~Yershov, and E.~Frazzoli, ``A survey of motion planning and control techniques for self-driving urban vehicles,'' \emph{IEEE Transactions on Intelligent Vehicles}, vol.~1, no.~1, 2016.

\bibitem{Teng2023}
S.~Teng and et~al., ``Motion planning for autonomous driving: The state of the art and future perspectives,'' \emph{IEEE Transactions on Intelligent Vehicles}, vol.~8, no.~6, 2023.

\bibitem{Zhou2022}
C.~Zhou, B.~Huang, and P.~Fr{\"a}nti, ``A review of motion planning algorithms for intelligent robots,'' \emph{Journal of Intelligent Manufacturing}, vol.~33, no.~2, 2022.

\bibitem{Gonzlez2015}
D.~González, J.~Pérez, V.~Milanés, and F.~Nashashibi, ``A review of motion planning techniques for automated vehicles,'' \emph{IEEE Transactions on Intelligent Transportation Systems}, vol.~17, no.~4, 2016.

\bibitem{Dong2023}
L.~Dong, Z.~He, C.~Song, and C.~Sun, ``A review of mobile robot motion planning methods: from classical motion planning workflows to reinforcement learning-based architectures,'' \emph{Journal of Systems Engineering and Electronics}, vol.~34, no.~2, 2023.

\bibitem{Tampuu2022}
A.~Tampuu, T.~Matiisen, M.~Semikin, D.~Fishman, and N.~Muhammad, ``A survey of end-to-end driving: Architectures and training methods,'' \emph{IEEE Transactions on Neural Networks and Learning Systems}, vol.~33, no.~4, 2022.

\bibitem{Frenetix}
R.~Trauth, K.~Moller, G.~Würsching, and J.~Betz, ``Frenetix: A high-performance and modular motion planning framework for autonomous driving,'' \emph{IEEE Access}, 2024.

\bibitem{nyberg2021risk}
T.~Nyberg, C.~Pek, L.~Dal~Col, C.~Nor{\'e}n, and J.~Tumova, ``Risk-aware motion planning for autonomous vehicles with safety specifications,'' in \emph{2021 IEEE intelligent vehicles symposium (IV)}.\hskip 1em plus 0.5em minus 0.4em\relax IEEE, 2021.

\bibitem{trauth2023toward}
R.~Trauth, K.~Moller, and J.~Betz, ``Toward safer autonomous vehicles: Occlusion-aware trajectory planning to minimize risky behavior,'' \emph{IEEE Open Journal of Intelligent Transportation Systems}, vol.~4, 2023.

\bibitem{Moller2024}
K.~Moller, R.~Trauth, and J.~Betz, ``Overcoming blind spots: Occlusion considerations for improved autonomous driving safety,'' in \emph{2024 IEEE Intelligent Vehicles Symposium (IV)}.\hskip 1em plus 0.5em minus 0.4em\relax IEEE, Jun. 2024.

\bibitem{Piazza2024}
M.~Piazza, M.~Piccinini, S.~Taddei, and F.~Biral, ``Mptree: A sampling-based vehicle motion planner for real-time obstacle avoidance,'' \emph{IFAC-PapersOnLine}, vol.~58, no.~10, p. 146–153, 2024.

\bibitem{Monderman2006}
H.~Monderman, E.~Clarke, and B.~H. Baillie, ``Shared space - the alternative approach to calming traffic,'' \emph{Traffic engineering and control}, vol.~47, pp. 290--292, 2006.

\bibitem{Liu2015}
W.~Liu, Z.~Weng, Z.~Chong, X.~Shen, S.~Pendleton, B.~Qin, G.~M.~J. Fu, and M.~H. Ang, ``Autonomous vehicle planning system design under perception limitation in pedestrian environment,'' in \emph{Conference on Cybernetics and Intelligent Systems (CIS) and IEEE Conference on Robotics, Automation and Mechatronics (RAM)}, 2015.

\bibitem{Morales2016}
N.~Morales, R.~Arnay, J.~Toledo, A.~Morell, and L.~Acosta, ``Safe and reliable navigation in crowded unstructured pedestrian areas,'' \emph{Engineering Applications of Artificial Intelligence}, vol.~49, p. 74–87, Mar. 2016.

\bibitem{Yang2023}
B.~Yang, S.~Yan, Z.~Wang, and K.~Nakano, ``Prediction based trajectory planning for safe interactions between autonomous vehicles and moving pedestrians in shared spaces,'' \emph{IEEE Transactions on Intelligent Transportation Systems}, vol.~24, no.~10, 2023.

\bibitem{Bai2015}
H.~Bai, S.~Cai, N.~Ye, D.~Hsu, and W.~S. Lee, ``Intention-aware online pomdp planning for autonomous driving in a crowd,'' in \emph{International Conference on Robotics and Automation (ICRA)}, 2015.

\bibitem{Luo2018}
Y.~Luo, P.~Cai, A.~Bera, D.~Hsu, W.~S. Lee, and D.~Manocha, ``Porca: Modeling and planning for autonomous driving among many pedestrians,'' \emph{IEEE Robotics and Automation Letters}, vol.~3, no.~4, 2018.

\bibitem{Li2024}
D.~Li, Y.~Jiang, J.~Zhang, and B.~Xiao, ``Smpc-based motion planning of automated vehicle when interacting with occluded pedestrians,'' \emph{IEEE Transactions on Intelligent Transportation Systems}, vol.~25, no.~12, 2024.

\bibitem{Zhu2022}
H.~Zhu, T.~Han, W.~K. Alhajyaseen, M.~Iryo-Asano, and H.~Nakamura, ``Can automated driving prevent crashes with distracted pedestrians? an exploration of motion planning at unsignalized mid-block crosswalks,'' \emph{Accident Analysis and Prevention}, vol. 173, 2022.

\bibitem{Li2020}
K.~Li, M.~Shan, K.~Narula, S.~Worrall, and E.~Nebot, ``Socially aware crowd navigation with multimodal pedestrian trajectory prediction for autonomous vehicles,'' in \emph{International Conference on Intelligent Transportation Systems (ITSC)}, 2020.

\bibitem{geisslinger2021autonomous}
M.~Geisslinger, F.~Poszler, J.~Betz, C.~L{\"u}tge, and M.~Lienkamp, ``Autonomous driving ethics: From trolley problem to ethics of risk,'' \emph{Philosophy \& Technology}, vol.~34, no.~4, pp. 1033--1055, 2021.

\bibitem{nuScenes}
H.~Caesar, V.~Bankiti, A.~H. Lang, S.~Vora, V.~E. Liong, Q.~Xu, A.~Krishnan, Y.~Pan, G.~Baldan, and O.~Beijbom, ``nuscenes: A multimodal dataset for autonomous driving,'' in \emph{2020 IEEE/CVF Conference on Computer Vision and Pattern Recognition (CVPR)}, 2020.

\bibitem{Dauner2023CORL}
D.~Dauner, M.~Hallgarten, A.~Geiger, and K.~Chitta, ``Parting with misconceptions about learning-based vehicle motion planning,'' in \emph{Conference on Robot Learning (CoRL)}, 2023.

\bibitem{caesar_nuplan_2022}
H.~Caesar, J.~Kabzan, K.~S. Tan, W.~K. Fong, E.~Wolff, A.~Lang, L.~Fletcher, O.~Beijbom, and S.~Omari, ``{NuPlan}: {A} closed-loop {ML}-based planning benchmark for autonomous vehicles,'' Feb. 2022.

\bibitem{idm}
M.~Treiber, A.~Hennecke, and D.~Helbing, ``Congested traffic states in empirical observations and microscopic simulations,'' \emph{Phys. Rev. E}, vol.~62, 2000.

\bibitem{hallgarten_can_2024}
M.~Hallgarten, J.~Zapata, M.~Stoll, K.~Renz, and A.~Zell, ``\BIBforeignlanguage{en}{Can {Vehicle} {Motion} {Planning} {Generalize} to {Realistic} {Long}-tail {Scenarios}?}'' Apr. 2024.

\bibitem{chitta_sledge_2024}
K.~Chitta, D.~Dauner, and A.~Geiger, ``\BIBforeignlanguage{en}{{SLEDGE}: {Synthesizing} {Simulation} {Environments} for {Driving} {Agents} with {Generative} {Models}},'' Mar. 2024.

\bibitem{helbing_social_1995}
D.~Helbing and P.~Molnár, ``\BIBforeignlanguage{en}{Social force model for pedestrian dynamics},'' \emph{\BIBforeignlanguage{en}{Physical Review E}}, vol.~51, no.~5, 1995.

\bibitem{zeng_modified_2014}
W.~Zeng, H.~Nakamura, and P.~Chen, ``A {Modified} {Social} {Force} {Model} for {Pedestrian} {Behavior} {Simulation} at {Signalized} {Crosswalks},'' \emph{Procedia - Social and Behavioral Sciences}, vol. 138, 2014.

\bibitem{johansson_waiting_2015}
F.~Johansson, A.~Peterson, and A.~Tapani, ``Waiting pedestrians in the social force model,'' \emph{Physica A: Statistical Mechanics and its Applications}, vol. 419, 2015.

\bibitem{kreiss_deep_2021}
S.~Kreiss, ``Deep {Social} {Force},'' Sep. 2021.

\bibitem{SUMO2018}
P.~A. Lopez, M.~Behrisch, L.~Bieker-Walz, J.~Erdmann, Y.-P. Fl{\"o}tter{\"o}d, R.~Hilbrich, L.~L{\"u}cken, J.~Rummel, P.~Wagner, and E.~Wie{\ss}ner, ``Microscopic traffic simulation using sumo,'' in \emph{The 21st IEEE International Conference on Intelligent Transportation Systems}, 2018.

\bibitem{chraibi_jupedsim_2024}
\BIBentryALTinterwordspacing
M.~Chraibi, K.~Kratz, T.~Schrödter, and {The JuPedSim Development Team}, ``{JuPedSim},'' May 2024. [Online]. Available: \url{https://github.com/PedestrianDynamics/jupedsim}
\BIBentrySTDinterwordspacing

\bibitem{klischat_coupling2019}
M.~Klischat, O.~Dragoi, M.~Eissa, and M.~Althoff, ``Coupling sumo with a motion planning framework for automated vehicles,'' in \emph{SUMO User Conference 2019}, vol.~62.\hskip 1em plus 0.5em minus 0.4em\relax EasyChair, 2019, pp. 1--9.

\bibitem{Corbetta2018}
A.~Corbetta, J.~A. Meeusen, C.-m. Lee, R.~Benzi, and F.~Toschi, ``Physics-based modeling and data representation of pairwise interactions among pedestrians,'' \emph{Physical Review E}, vol.~98, no.~6, Dec. 2018.

\bibitem{gao_yuxiang-gaopysocialforce_2024}
\BIBentryALTinterwordspacing
Y.~Gao, ``yuxiang-gao/{PySocialForce},'' Jun. 2024. [Online]. Available: \url{https://github.com/yuxiang-gao/PySocialForce}
\BIBentrySTDinterwordspacing

\bibitem{noauthor_srl-freiburgpedsim_ros_2024}
\BIBentryALTinterwordspacing
``srl-freiburg/pedsim\_ros,'' Jun. 2024. [Online]. Available: \url{https://github.com/srl-freiburg/pedsim_ros}
\BIBentrySTDinterwordspacing

\bibitem{thrun2005probabilistic}
S.~Thrun, W.~Burgard, and D.~Fox, \emph{Probabilistic Robotics}, ser. Intelligent Robotics and Autonomous Agents series.\hskip 1em plus 0.5em minus 0.4em\relax MIT Press, 2005.

\bibitem{Shiffrin2012}
S.~V. Shiffrin, ``{HARM} {AND} {ITS} {MORAL} {SIGNIFICANCE},'' \emph{Legal Theory}, vol.~18, no.~3, Aug. 2012.

\bibitem{BundesministeriumVerkehr2017}
\BIBentryALTinterwordspacing
{Ethik-Kommission Automatisiertes und Vernetztes Fahren}, ``Automatisiertes und vernetztes fahren,'' 2017. [Online]. Available: \url{https://bmdv.bund.de/SharedDocs/DE/Publikationen/DG/bericht-der-ethik-kommission.pdf}
\BIBentrySTDinterwordspacing

\bibitem{gennarelli2006ais}
T.~A. Gennarelli and E.~Wodzin, ``Ais 2005: a contemporary injury scale,'' \emph{Injury}, vol.~37, no.~12, 2006.

\bibitem{Kleinbaum2010}
D.~G. Kleinbaum and M.~Klein, ``Logistic regression,'' \emph{Statistics for Biology and Health}, 2010.

\bibitem{NHTSA2024}
\BIBentryALTinterwordspacing
{National Highway Traffic Safety Administration}. (2024) Crash report sampling system: Motor vehicle crash data collection. Washington, DC. [Online]. Available: \url{https://www.nhtsa.gov/crash-data-systems/crash-report-sampling-system}
\BIBentrySTDinterwordspacing

\bibitem{walenet}
M.~Geisslinger, P.~Karle, J.~Betz, and M.~Lienkamp, ``Watch-and-learn-net: Self-supervised online learning for probabilistic vehicle trajectory prediction,'' in \emph{2021 IEEE International Conference on Systems, Man, and Cybernetics (SMC)}, 2021.

\bibitem{Henze2024}
N.~Henze, \emph{Asymptotic Stochastics: An Introduction with a View towards Statistics}.\hskip 1em plus 0.5em minus 0.4em\relax Springer Berlin Heidelberg, 2024.

\bibitem{Lambert2008}
A.~Lambert, D.~Gruyer, and G.~Saint~Pierre, ``A fast monte carlo algorithm for collision probability estimation,'' in \emph{2008 10th International Conference on Control, Automation, Robotics and Vision}, 2008.

\end{thebibliography}
